\pdfoutput=1

\documentclass[11pt]{article}

\usepackage[final]{acl}

\usepackage{times}
\usepackage{latexsym}

\usepackage[T1]{fontenc}

\usepackage[utf8]{inputenc}

\usepackage{microtype}

\usepackage{inconsolata}

\usepackage{graphicx}

\usepackage{booktabs}
\usepackage{CJKutf8}
\usepackage{multirow}
\usepackage{multicol}
\usepackage{amsfonts}
\usepackage{amsmath}
\usepackage{tcolorbox} 
\usepackage{subcaption}
\usepackage{enumitem}

\usepackage{xcolor,colortbl}
\definecolor{Gray}{gray}{0.90}
\newcolumntype{a}{>{\columncolor{Gray}}c}

\everypar{\looseness=-1}

\title{Understanding and Mitigating Language Confusion in LLMs}

\newcommand*\samethanks[1][\value{footnote}]{\footnotemark[#1]}

\author{
    \textbf{Kelly Marchisio}\thanks{Equal contribution. Correspondence: \texttt{sebastianruder@cohere.com}.},
    \textbf{Wei-Yin Ko}\samethanks,
    \textbf{Alexandre Bérard},
    \textbf{Théo Dehaze},
    \textbf{Sebastian Ruder}\samethanks
\\
    Cohere
}

\begin{document}

\maketitle

\begin{abstract}
We investigate a surprising limitation of LLMs: their inability to consistently generate text in a user's desired language. We create the Language Confusion Benchmark (LCB) to evaluate such failures, covering 15 typologically diverse languages with existing and newly-created English and multilingual prompts. We evaluate a range of LLMs on monolingual and cross-lingual generation reflecting practical use cases, finding that Llama Instruct and Mistral models exhibit high degrees of language confusion and even the strongest models fail to consistently respond in the correct language. 
We observe that base and English-centric instruct models are more prone to language confusion, which is aggravated by 
complex prompts and high sampling temperatures. We find that language confusion can be partially mitigated via few-shot prompting, multilingual SFT and preference tuning. We release our language confusion benchmark, which serves as a first layer of efficient, scalable multilingual evaluation.\footnote{The data is available at \url{https://github.com/for-ai/language-confusion}.}
\end{abstract}

\section{Introduction}

Large language models (LLMs) are increasingly used in a variety of applications across the globe \cite{kaddour2023challenges}.
While early LLMs focused on English \cite{joshi-etal-2020-state,hu2020xtreme}, recent models are more multilingual. Nevertheless, LLMs do not provide equal utility to non-English speakers due to higher latency, increased costs, and reduced performance \cite{ahia-etal-2023-languages,asai2023buffet,held2023material}.

\begin{figure}[t]
\centering
\includegraphics[scale=0.4]{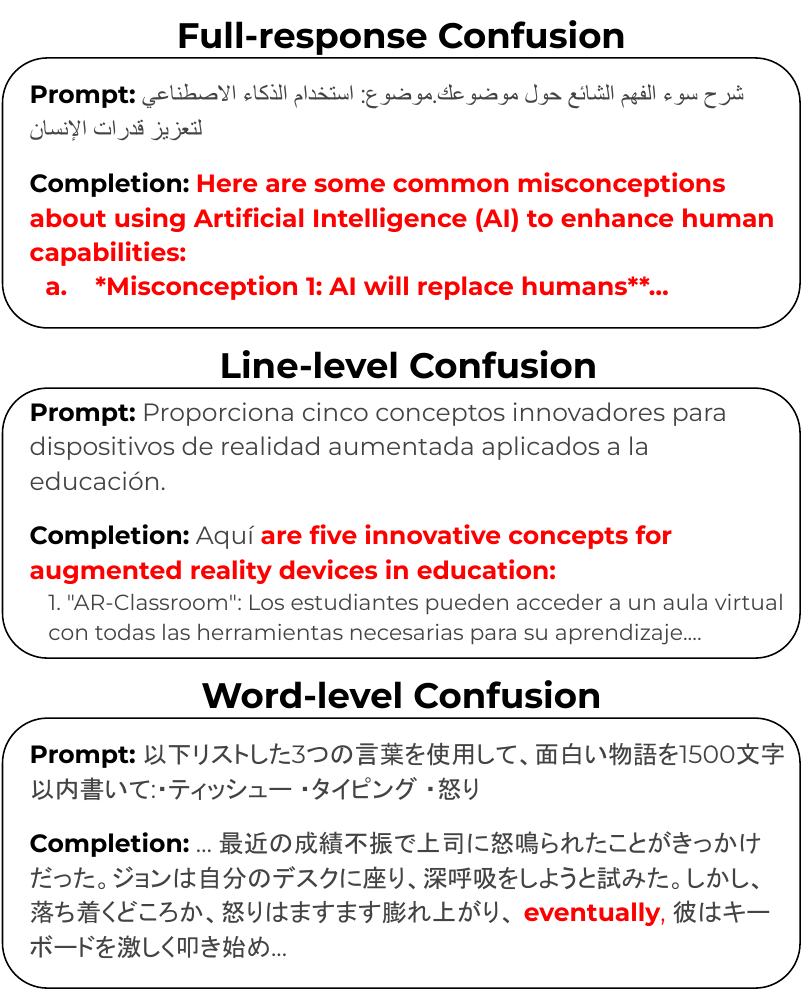}
\caption{Language Confusion can occur at the word level, line level, or over the entire output response.}
\label{fig:example}
\end{figure}
To be useful, an LLM must understand intent and provide a response that is appropriate in both form, e.g., correct grammar, style, tone, register, and content, e.g., truthful, coherent, concise \cite{grice1975logic,wilson2012linguistic}.
While content-related issues such as hallucinations have attracted substantial attention \cite{ji2023towards,bang-etal-2023-multitask}, they are often subtle and difficult to evaluate \cite{gudibande2024false,hosking2024human}, particularly in multilingual settings \cite{guerreiro2023hallucinations}. Form-related errors indicate a more obvious failure to fulfill a request and---in extreme cases---may cause confusing or unintelligible responses.

We identify a surprising form limitation that drastically reduces LLMs' utility for non-English languages: LLMs are often unable to consistently generate text in the user's desired language, or the appropriate language given the context. We call this category of error ``language confusion''.\footnote{We use this term to indicate that this is an \emph{erroneous} behaviour rather than \emph{natural} alternations between languages, i.e., code-switching \cite{dogruoz-etal-2021-survey}.}

Take an Arabic prompt as an example: an LLM may inappropriately respond fully in English (full-response confusion), produce some lines in the desired language and some in another language (line-level confusion), or sporadically insert single words or phrases in another language (word-level confusion). Figure~\ref{fig:example} shows example errors.  Even if such errors occur rarely, they cause a jarring user \begin{CJK}{UTF8}{gbsn}经验\end{CJK} (experience).


We investigate language confusion on the line and word level
in two practical settings: \textbf{a)} Monolingual generation, where a user queries the LLM in a given language, \emph{implicitly} requesting an answer in the same language; and \textbf{b)} cross-lingual generation, where a user \emph{explicitly} instructs a model to generate text in a different language.

We create and release a language confusion benchmark covering 15 typologically diverse languages, sourcing prompts from public English and multilingual instruction datasets, and additionally creating new data with more complex prompts. We evaluate a range of state-of-the-art LLMs including Llama, Command R, Mistral, and OpenAI family models. We find that Llama Instruct and Mistral LLMs exhibit severe language confusion in many languages. While Command R and OpenAI models fare much better on monolingual generation, even the strongest cannot consistently generate text in the correct language cross-lingually.

\vspace{3pt}
\noindent Our contributions are the following: 
\setlist{nolistsep}
\begin{enumerate}[nolistsep]
    \item We identify and describe the issue of language confusion in LLMs.
    \item We introduce a new benchmark and metrics to measure language confusion in LLMs.
    \item We perform a systematic evaluation of various LLMs, investigating when language confusion occurs in practice.
    \item We propose methods to mitigate language confusion in LLMs.
\end{enumerate}


\begin{table*}[h]
\centering
\resizebox{\textwidth}{!}{%
\begin{tabular}{llllcclc}
\toprule
 & Dataset name & Reference & Nature of data & $|L|$ & $|D|$ & Languages & $W$ \\ \midrule
\multirow{5}{*}{\rotatebox[origin=c]{90}{\parbox[c]{1.5cm}{\centering Mono-lingual}}} & Aya & \citet{singh2024aya} & Human-generated & 100 & 500 & en, tr, ar, zh, pt & 9 \\
 & Dolly & \citet{singh2024aya} & MT post-edited & 100 & 500 & hi, ru, fr, ar, es & 10 \\
  \addlinespace[3pt]
 & \multirow{2}{*}{Okapi} & \multirow{2}{*}{\citet{lai-etal-2023-okapi}} & \multirow{2}{*}{Synthetic + MT} & \multirow{2}{*}{100} & \multirow{2}{*}{1.2k} & en, fr, it, de, zh, & \multirow{2}{*}{13} \\
 &  &  &  &  &  & vi, es, id, pt, ar \\
 \addlinespace[3pt]
 & Native prompts & Ours & Human-generated & 100 & 400 & es, fr, ja, ko & 19 \\ \midrule
\multirow{2}{*}{\rotatebox[origin=c]{90}{\parbox[c]{1.5cm}{\centering Cross-lingual}}} & Okapi & \citet{lai-etal-2023-okapi} & Synthetic & 100 & 1.5k & $\mathcal{L}$ & 15 \\
 & ShareGPT & \url{https://sharegpt.com/} & Human-generated & 100 & 1.5k & $\mathcal{L}$ & 18 \\
& Complex prompts & Ours & Human-generated & 99 & 1.5k & $\mathcal{L}$ & 159 \\
\bottomrule
\end{tabular}%
}
\caption{\textbf{Data sources in the LCB for monolingual and cross-lingual generation.} $|D|$ is the total number of examples per data source and $|L|$ is the number of examples per language. For the cross-lingual setting, the model is instructed in English to generate in the target language $l \in \mathcal{L}$ where $\mathcal{L} = \{\text{fr, de, es, pt, it, ja, ko, zh, ar, tr, hi, ru, id, vi} \}$. $W$ is the median length in words of the prompts in each dataset.}
\label{tab:dataset-stats}
\end{table*}

\section{Language Confusion Benchmark}

While some datasets to evaluate LLMs' performance on \emph{natural} code-switched data exist \cite{khanuja-etal-2020-gluecos,winata-etal-2023-decades}, there are none designed to assess language confusion in LLMs. We create the Language Confusion Benchmark (LCB) by collecting a diverse set of prompts reflecting realistic use cases across a typologically diverse set of languages. The benchmark is easily extensible, cheap, and efficient to evaluate.

\subsection{Generation settings}

We measure language confusion in two settings: monolingual and cross-lingual generation.

\paragraph{Monolingual generation} A speaker queries a model in language $l$ and expects a response in $l$. This is the most common usage scenario as users often prefer to interact with technology in their native language \cite{kantar_internet_2023}.


\paragraph{Cross-lingual generation} A user instructs a model in language $l$ to fulfill a request in another language $l'$. In this challenging setting, the requested language $l'$ is \textit{different} from the instruction language $l$. This setting is relevant in applications where multilingual outputs are required, but optimizing a prompt for each input language is inefficient in practice or when a user requires a generation in a language they do not speak. We set the instruction language to English.

\subsection{Language Confusion Metrics}
To detect language confusion, we rely on off-the-shelf language identification (LID) tools. We employ fastText \cite{joulin2016bag} as a strong alternative to expensive LLM-based evaluation. 

\paragraph{Line-level detection} We split a response into lines (by newline character) and check each line against the user's desired language with fastText.\footnote{We only apply fastText to sequences of more than 
4 words as its LID predictions are less precise for shorter sequences.}

\paragraph{Word-level detection} Off-the-shelf tools do not support word-level LID and LLMs only achieve 79--86 F1 detecting word-level code-switching \cite{zhang-etal-2023-multilingual}, too low for use as automatic evaluators. Consequently, we take a two-pronged heuristic approach to detecting word-level language confusion focusing on settings where it achieves high precision. For non-Latin script languages, we observe that word-level language confusion in top LLMs mainly occurs with English. 
To avoid natural code-switching false positives,\footnote{E.g., Japanese users may use English acronyms: \begin{CJK}{UTF8}{min}AI に関する記事を書きます。\end{CJK}(``Write an article about AI.'')} we check for English words that do not typically occur in target language text.\footnote{We source words from \url{https://gist.github.com/WChargin/8927565}, which is based on the Linux dictionary usually stored in \texttt{/usr/share/dict/words}. To reduce false positives, we exclude capitalized words from the dictionary (which are often proper nouns or acronyms).} We evaluate word-level confusion in Arabic (ar), Hindi (hi), Japanese (ja), Korean (ko), Russian (ru), and Simplified Chinese (zh). For Latin script languages where word-level language confusion is rarer,\footnote{We show an example of such word-level confusion in \ref{fig:non_en_word_switch}.} we detect tokens where any character is outside of the Unicode range of the language's script. We evaluate word-level confusion in German (de), English (en), Spanish (es), French (fr), Indonesian (id), Italian (it), Portuguese (pt), Turkish (tr), and Vietnamese (vi).

\paragraph{Binary evaluation} A response is only correct when entirely in the correct language, as even one instance of language confusion can damage intelligibility and cause a jarring user experience.  We calculate binary metrics to indicate whether a response contains any instance of \textbf{a)} a line in an incorrect language and \textbf{b)} an isolated English word for languages using $\neg$ Latin scripts and an out-of-Unicode-range character for Latin script languages. These main metrics are defined below.

\textit{Line-level pass rate} (\textbf{LPR}): percentage of model responses that pass our line-level language confusion detector without error. A response is ``correct'' if all lines match the user's desired language.

$$
\text{LPR} = \frac{|R \setminus E_L|}{|R|}
$$
where $R$ is the set of all responses and $E_L$ the set of responses that contain line-level errors.\footnote{For completeness, we show results for ``line-level language accuracy'' (the fraction of lines in the correct language across all responses of an LLM) in Tables \ref{tab:monolingual-line-acc} and \ref{tab:crosslingual-line-acc} at \S\ref{sec:extended_results}.}

\textit{Word-level pass rate} (\textbf{WPR}): percentage of responses where all words are in the desired language. We exclude responses with line-level errors as most line-level errors would also be counted as word-level errors, making it difficult to disentangle the two error types.
For languages that use a $\neg$ Latin script, we detect erroneous English words while for Latin script languages, we identify characters outside of the script's Unicode range.


$$
\text{WPR} = \frac{|(R \setminus E_L) \setminus E_W|}{|R \setminus E_L|}
$$
where $R$ is the set of all responses, $E_L$ the set of responses with line-level errors and $E_W$ the set of responses with word-level errors.

\begin{table*}[h]
\centering
\small
\begin{tabular}{lp{70mm}p{60mm}p{9mm}}
\toprule
 & \multirow{2}{*}{Prompt} & \multirow{2}{*}{Translation} & Data source \\
 \midrule
\multirow{6}{*}{\rotatebox[origin=c]{90}{Monolingual}} & \begin{CJK*}{UTF8}{gbsn}问：如何清洗和保养筷子？\end{CJK*} & Q: How to clean and maintain chopsticks? & Aya \\[1.5ex]  
 & ¿Cómo escapar de un helicóptero atrapado en el agua? & How to escape from a helicopter stuck in water? & Dolly \\[1.5ex] 
  & Erklären Sie, wie der Gini-Index berechnet wird. & Explain how the Gini index is calculated. & Okapi \\[1.5ex] 
 & \begin{CJK}{UTF8}{min}日本で救急隊員を目指す人は、どのような教育やトレーニングを受ける必要がありますか。\end{CJK} & What kind of education and training do people who want to become emergency workers in Japan need to undergo? & Native prompts (Ours) \\ \midrule
\multirow{7}{*}{\rotatebox[origin=c]{90}{Cross-lingual}} & \multicolumn{2}{p{135mm}}{Generate an essay \textbf{in Korean} of at least 500 words that argues in favor of regulating artificial intelligence.\vspace{2mm}} & Okapi \\
\vspace{2mm}   & \multicolumn{2}{p{135mm}}{\textbf{Respond in French}. You are a medical communications expert. Please provide a summary on how pharma companies are approaching diversity and inclusion, and health inequalities globally. Focus on the general approach and include information on clinical trials.} & Share- GPT \\
 & \multicolumn{2}{p{135mm}}{Based solely on the text below: 1. Extract the statistical techniques and machine learning algorithms analysts employ to uncover relationships and patterns within the data. 2. Generate 5 fill-in-the-blanks style questions 3. Summarize the text in 100 words {[}...{]} \textbf{Reply in Turkish.}} & Complex prompts (Ours) \\
\bottomrule
\end{tabular}%
\caption{\textbf{An example prompt from each dataset used for monolingual and cross-lingual generation.} English translations are shown for convenience. For cross-lingual generation, prompts are in English and have been amended with an instruction to generate the output in another language. The complex prompt example is truncated.}
\label{tab:example-prompts}
\end{table*}

\textit{Language confusion pass rate} (\textbf{LCPR}): harmonic mean\footnote{Similar to F1, this gives higher importance to low values.} of LPR and WPR:
\[
\text{LCPR} = 2\times\frac{\text{LPR} \times \text{WPR}}{\text{LPR} + \text{WPR}}
\]
LCPR is elucidating in cases of severe issues producing output in the correct language.\footnote{Imagine a model's LPR for Arabic is a dismal 1\% because only 1/100 responses are fully Arabic, the rest English. WPR is a deceptively high 100\%: as outputs are fully English, 
no English word appears in an Arabic line.  
The model scores 2\% LCPR, however, reflective of severe confusion.
} 

\subsection{Data sources}

The monolingual and cross-lingual tasks respectively comprise 2600 and 4500 prompts in total, across 15 typologically diverse languages: English, French, German, Spanish, Portuguese, Italian, Japanese, Korean, Chinese, Arabic, Turkish, Hindi, Russian, Indonesian, and Vietnamese. Details are shown in Table \ref{tab:dataset-stats}.

The prompts are sourced from the datasets below, focusing on human-annotated or human-edited prompts. We filter each dataset to make it most useful for evaluating language confusion.

\paragraph{Aya} 250 original human-written prompts in 7 languages each from the \texttt{aya-human-annotated} subset of the Aya Evaluation Suite \cite{singh2024aya}.

\paragraph{Dolly} 200 machine-translated Dolly \cite{conover2023free} prompts post-edited by fluent speakers for 6 languages from the Aya Evaluation Suite's \texttt{dolly-human-edited} subset \cite{singh2024aya}.

\paragraph{Okapi} We use the \texttt{multilingual-alpaca-52k} subset of the Okapi data, which contains ChatGPT-generated translations of 52k English instructions from Alpaca \cite{taori2023alpaca} into 26 languages.

\paragraph{ShareGPT} We use prompts from the first turn of 90,000 mostly English user conversations with ChatGPT, scraped via the ShareGPT API\footnote{\url{https://sharegpt.com/}} before it was shut down.\footnote{\url{https://huggingface.co/datasets/RyokoAI/ShareGPT52K}}

\paragraph{Native prompts (Ours)} For Japanese and Korean, under-represented in the above datasets, as well as Spanish and French, we commission native annotators to collect our own prompts (see \S\ref{sec:app-annotator-stats}).

\paragraph{Complex prompts (Ours)} As prompts from the above sources are relatively simple, we collect complex English prompts written by human annotators.

\subsection{Data Filtering and Processing}

\paragraph{Suitability for LID} As LID tools underperform on short sequences and non-standard text, we manually filter out: \textbf{a)} examples answerable with a single word/phrase; \textbf{b)} multiple-choice questions and prompts asking for lists; \textbf{c)} prompts requiring code generation, math equations, or data formats such as HTML. For datasets where completions are available, we filter out prompts with very short completions (less than 5 words). 

\paragraph{Western-centric responses} Many datasets created via translation contain questions about Western-centric concepts (e.g., US National Parks, presidents or US-based brands) which can cause false positives with our word-level detector. We manually filter out such questions.

\paragraph{Prompt format} For cross-lingual generation, we semi-automatically amend prompts with an instruction to generate in a target language (see \S\ref{sec:app-cross-lingual-prompt-generation} for details). Prompts are used as-is for monolingual generation. Some examples are shown in Table \ref{tab:example-prompts}.

\begin{table*}[h]
\centering
\resizebox{\textwidth}{!}{%
\begin{tabular}{@{}laccccccccccccccc@{}}
\toprule
\multicolumn{1}{l}{} & \multicolumn{16}{c}{\textbf{Monolingual}}                                                        \\
\multicolumn{1}{l}{} & avg           & ar             & de             & en             & es            & fr             & hi             & id            & it             & ja             & ko             & pt            & ru             & tr             & vi             & zh            \\  \midrule
Llama 2 70B-I        & 48.3          & 0.3            & 59.0           & 99.0           & 95.7          & 87.7           & 1.0            & 62.0          & 72.0           & 7.0            & 0.0            & 91.0          & 88.9           & 33.0           & 17.0           & 10.5          \\
Llama 3 70B-I        & 46.0          & 21.7           & 31.0           & \textbf{100.0} & 98.3          & 88.7           & 23.0           & 21.0          & 88.0           & 10.0           & 0.0            & 95.5          & 77.0           & 18.0           & 10.0           & 8.0           \\
Llama 3.1 70B-I & 99.0 & 98.9 & \textbf{100.0} & 98.5 & 99.0 & \textbf{100.0} & \textbf{100.0} & 94.0 & \textbf{100.0} & 96.9 & \textbf{100.0} & \textbf{99.0} & \textbf{100.0} & \textbf{100.0} & \textbf{100.0} & \textbf{99.0} \\
Mixtral 8x7B         & 73.0          & 48.3           & 90.9           & 99.5           & 89.3          & 95.3           & 71.0           & 58.0          & 72.0           & 66.7           & 61.2           & 85.0          & 65.0           & 90.0           & 57.0           & 45.5          \\
Mistral Large        & 69.9          & 48.0           & 98.0           & 99.0           & 99.0          & \textbf{100.0} & 19.0           & 31.0          & 99.0           & 48.0           & 64.0           & 79.5          & 98.0           & 71.0           & 29.0           & 66.0          \\
Command R            & 98.6          & \textbf{100.0} & 98.0           & 99.5           & 95.7          & 99.3           & \textbf{100.0} & 92.0          & 99.0           & \textbf{100.0} & \textbf{100.0} & 98.5 & \textbf{100.0} & 99.0           & 99.0           & 98.5          \\
Command R+           & 99.2          & 99.7           & \textbf{100.0} & \textbf{100.0} & 99.3          & 99.7           & \textbf{100.0} & \textbf{97.0} & \textbf{100.0} & 99.0           & \textbf{100.0} & 97.5          & \textbf{100.0} & \textbf{100.0} & 99.0           & 97.5          \\
Command R Refresh & 98.9 & 99.6 & \textbf{100.0} & 99.5 & 99.3 & 99.7 & \textbf{100.0} & 92.0 & \textbf{100.0} & 99.0 & \textbf{100.0} & 98.0 & \textbf{100.0} & 99.0 & \textbf{100.0} & 98.0 \\
Command R+ Refresh & \textbf{99.3} & 99.0 & \textbf{100.0} & \textbf{100.0} & 99.3 & \textbf{100.0} & \textbf{100.0} & 96.0 & \textbf{100.0} & \textbf{100.0} & \textbf{100.0} & 97.5 & 99.0 & \textbf{100.0} & \textbf{100.0} & 98.0 \\
GPT-3.5 Turbo        & 99.1          & \textbf{100.0} & \textbf{100.0} & 99.5           & \textbf{99.7} & \textbf{100.0} & 99.0           & 96.0          & \textbf{100.0} & 98.0           & \textbf{100.0} & 98.0          & \textbf{100.0} & \textbf{100.0} & 99.0           & 97.0          \\
GPT-4 Turbo          & \textbf{99.3} & 99.0           & \textbf{100.0} & \textbf{100.0} & 99.3          & 99.3           & \textbf{100.0} & 96.0          & 99.0           & \textbf{100.0} & \textbf{100.0} & 98.0          & \textbf{100.0} & \textbf{100.0} & \textbf{100.0} & \textbf{99.0} \\
GPT-4o & 98.9 & 99.7 & \textbf{100.0} & \textbf{100.0} & 99.3 & 99.3 & 99.0 & 94.0 & \textbf{100.0} & 99.0 & \textbf{100.0} & 97.5 & 99.0 & \textbf{100.0} & 99.0 & 98.0 \\ \midrule
\multicolumn{1}{l}{} & \multicolumn{16}{c}{\textbf{Cross-lingual}} \\ 
\multicolumn{1}{l}{} & avg           & ar             & de             & -              & es            & fr             & hi             & id            & it             & ja             & ko             & pt            & ru             & tr             & vi             & zh            \\ \midrule
Llama 2 70B-I        & 38.4          & 12.4           & 52.3           & -              & 77.3          & 71.1           & 21.2           & 46.0          & 66.5           & 16.2           & 4.8            & 75.9          & 38.3           & 24.0           & 20.4           & 11.1          \\
Llama 3 70B-I        & 30.3          & 31.1           & 34.7           & -              & 61.1          & 53.1           & 46.4           & 25.4          & 36.4           & 1.4            & 0.8            & 54.4          & 38.4           & 17.4           & 18.7           & 4.3           \\
Llama 3.1 70B-I & 81.4 & 77.2 & 87.5 & & 90.4 & 90.5 & 95.6 & \textbf{97.1} & 88.1 & 59.4 & 51.5 & 86.0 & 76.5 & 85.7 & 93.6 & 69.9 \\
Mixtral 8x7B         & 69.0          & 59.1           & 76.4           & -              & 79.1          & 79.0           & 39.2           & 72.8          & 85.0           & 57.9           & 56.9           & 79.4          & 72.4           & 76.0           & 75.8           & 57.5          \\
Mistral Large        & 58.2          & 36.1           & 74.5           & -              & 68.5          & 71.9           & 58.5           & 59.2          & 65.8           & 44.5           & 41.1           & 64.5          & 63.3           & 65.9           & 54.8           & 46.5          \\
Command R            & 68.1          & 61.6           & 63.2           & -              & 72.5          & 74.4           & 65.5           & 70.8          & 65.7           & 65.3           & 69.2           & 67.2          & 69.4           & 67.7           & 65.7           & 75.0          \\
Command R+           & 91.2 & 93.4  & 91.6           & -              & 91.7          & 91.5  & 90.2           & 85.9          & 93.8  & 93.8  & 91.1  & 88.5          & 93.0  & 92.0  & 91.1           & 89.5 \\
Command R Refresh & 93.1 & 91.9 & 96.1 & - & 96.4 & 94.0 & 95.0 & 85.1 & 93.8 & 95.0 & 93.8 & \textbf{94.0} & 92.2 & 93.4 & 94.1 & 88.9 \\
Command R+ Refresh & \textbf{95.4} & \textbf{95.4} & \textbf{97.5} & - & \textbf{97.6} & \textbf{97.2} & \textbf{98.2} & 88.9 & \textbf{96.2} & \textbf{95.1} & \textbf{95.9} & 91.7 & \textbf{96.4} & \textbf{97.9} & \textbf{97.9} & 90.2 \\
GPT-3.5 Turbo        & 89.8          & 90.8           & 90.2           & -              & 93.3 & 87.8           & 92.0  & 84.5          & 91.3           & 88.3           & 90.3           & 89.9          & 91.8           & 89.2           & 91.8  & 86.4          \\
GPT-4 Turbo          & 90.3          & 88.9           & 93.0  & -              & 93.1          & 90.7           & 91.0           & 87.3 & 91.8           & 87.7           & 89.7           & 91.0 & 90.0           & 91.4           & 90.0           & 87.9          \\
GPT-4o & 92.4 & 95.0 & 92.9 & - & 95.8 & 93.5 & 91.9 & 85.4 & 94.1 & 92.5 & 92.4 & 88.0 & 92.6 & 95.1 & 92.7 & \textbf{91.3} \\
\bottomrule
\end{tabular}
}
\caption{\textbf{Line-level pass rate (LPR) on monolingual and cross-lingual generation, by language.}
}
\label{tab:lang-lpr}
\end{table*}

\begin{table}[h]
\small
\centering
\begin{tabular}{@{}laa}
\toprule
\multicolumn{1}{l}{} & Monolingual & Cross-lingual \\ \midrule
Llama 2 70B-I        & 97.9            & 84.2              \\
Llama 3 70B-I        & 93.0            & 94.4              \\
Llama 3.1 70B-I      & 99.5            & 95.0              \\
Mixtral 8x7B         & 73.7            & 68.2              \\
Mistral Large        & 98.4            & 93.8              \\
Command R            & 96.3            & 94.0              \\
Command R+           & 99.4            & 95.1              \\
Command R Refresh    & 99.4            & 97.2              \\
Command R+ Refresh   & \textbf{99.8}   & 96.5              \\
GPT-3.5 Turbo        & \textbf{99.8}   & \textbf{98.7}     \\
GPT-4 Turbo          & 99.7            & 96.6              \\ 
GPT-4o               & 99.7            & 98.1              \\ 

\bottomrule
\end{tabular}%
\caption{\textbf{Average word-level pass rate (WPR) on non-Latin script languages}. See Tables \ref{tab:lang-word-level} and \ref{tab:lang-word-level-latin-script} for detailed WPR results on non-Latin and Latin script languages respectively.}
\label{tab:lang-wpr-avg}
\end{table}

\section{Experiments}
\label{sec:experiments}

\paragraph{Models} We evaluate the following LLMs covering various scales and model families: Llama 2 70B Instruct \cite{touvron2023llama}, Llama 3 and 3.1 70B Instruct \cite{dubey2024llama}, Command R (35B),\footnote{\url{https://cohere.com/blog/command-r}} Command R+ (104B parameters),\footnote{\url{https://cohere.com/blog/command-r-plus-microsoft-azure}} Command R Refresh (\texttt{command-r-08-2024}), Command R+ Refresh (\texttt{command-r-plus-08-2024}), Mixtral 8x7B \cite{jiang2024mixtral}, Mistral Large,\footnote{\url{https://mistral.ai/news/mistral-large/}} GPT-3.5 Turbo \cite[\texttt{gpt-3.5-turbo-012524};][]{brown2020language}, GPT-4 Turbo \cite[\texttt{gpt-4-turbo-040924};][]{achiam2023gpt}, and GPT-4o (\texttt{gpt-4o-2024-08-06}). For Llama and Command models, we also evaluate base versions (see \S\ref{sec:impact-instruction-tuning}).
We generate at most 100 tokens per prompt using nucleus sampling with $p=0.75$ and $T=0.3$.

LPR, WPR, and LCPR results are in Tables \ref{tab:lang-lpr},  \ref{tab:lang-wpr-avg}, and \ref{tab:lang-lcpr}, respectively.\footnote{We find results show little variance across runs (see \S\ref{sec:app-metric-variability}).} We show language confusion examples for different models in Table \ref{fig:lang-confuse-examples}.

\paragraph{Monolingual generation} Command and GPT models perform well on average on the line level (LPR in $[98.6, 99.3]$), 
but Llama 2 and 3 and Mistral models struggle to consistently generate text in the correct language (LPR in $[48.3, 73.0]$). Llama 3.1 performs much better, however.
Mistral Large is better on some European languages while Llama 2 and 3 exhibit language confusion even for high-resource languages such as German. Most models have WPR within the same range and perform better on Latin script vs non-Latin script languages; however, Mixtral 8x7B is considerably worse. GPT-4 Turbo is strongest on LCPR, with Command R+ Refresh and GPT-4o only slightly behind. Llama models' preference for English responses leads to very low LCPR.

\paragraph{Cross-lingual generation} In the challenging cross-lingual setting, the best models have LPRs in the low 90s. The Command R and R+ Refresh models outperform their original versions, with Command R Refresh in particular showing a large improvement over Command R, which performs poorly cross-lingually. OpenAI and Command models perform best; Command R+ Refresh achieves the best performance overall. Llama 2 and 3 models perform poorly: both scoring in the 30s due to a tendency to respond in English. Llama 3.1 shows improved performance but still tends to generate in English for some languages. Mistral Large is worse than Mixtral, even in European languages. On LCPR, Command R+ Refresh performs best, followed by GPT-4o.

\section{Analyses}
\label{sec:analyses}

\subsection{Impact of dataset} In creating the language confusion test sets, we aimed to include prompts covering various use cases and domains. We show differences of LPR by dataset in Tables~\ref{tab:dataset-impact-lpr} and \ref{tab:dataset-impact-lpr-monolingual} for cross-lingual and monolingual generation, with WPR and LCPR in Tables~\ref{tab:dataset-impact-wpr} and \ref{tab:dataset-impact-lcpr} in \S\ref{sec:dataset_lang_confuse}. Differences between datasets are small for monolingual generation.
Cross-lingually, the difference is more pronounced: models perform fairly well on Okapi and ShareGPT, but are much worse on our Complex prompts, indicating their more challenging nature.

\begin{table}[h]
\centering
\resizebox{\linewidth}{!}{%
\begin{tabular}{@{}la >{\centering\arraybackslash}p{1.4cm}>{\centering\arraybackslash}p{1.4cm}>{\centering\arraybackslash}p{1.4cm}@{}}
\toprule
\multicolumn{1}{l}{} & & \multirow{2}{*}{Okapi}         & \multirow{2}{*}{ShareGPT}      & Complex \\ 
& \multirow{-2}{*}{avg} &  & & (Ours) \\
\midrule
Llama 2 70B-I        & 38.4          & 43.4          & 46.0          & 25.7           \\
Llama 3 70B-I        & 30.3          & 35.0          & 39.5          & 16.3           \\
Llama 3.1 70B-I      & 81.3          & 85.3          & 91.0          & 67.7           \\
Mixtral 8x7B         & 69.0          & 77.4          & 79.8          & 49.9           \\
Mistral Large        & 58.2          & 68.0          & 56.2          & 50.5           \\
Command R            & 68.1          & 75.6          & 89.7          & 39.0           \\
Command R+           & 91.2 & 96.1          & 98.7 & 78.9  \\
Command R Refresh  & 93.1 & \textbf{98.3}  & \textbf{99.4}     & 81.7           \\
Command R+ Refresh & \textbf{95.4} & 98.0  & 98.7     & \textbf{89.6}           \\
GPT-3.5 Turbo        & 89.8          & 97.7 & 96.8          & 75.0           \\
GPT-4 Turbo          & 90.3          & 96.6          & 96.4          & 77.7           \\ 
GPT-4o             & 92.4 & 97.4  & 97.6     & 82.1         \\
\bottomrule
\end{tabular}
}
\caption{\textbf{Line-level pass rate (LPR) by dataset on cross-lingual generation.}}
\label{tab:dataset-impact-lpr}
\end{table}

\begin{figure*}[t]
    \centering
    \includegraphics[trim={0 0 0 2.61cm},clip, width=0.9\linewidth]{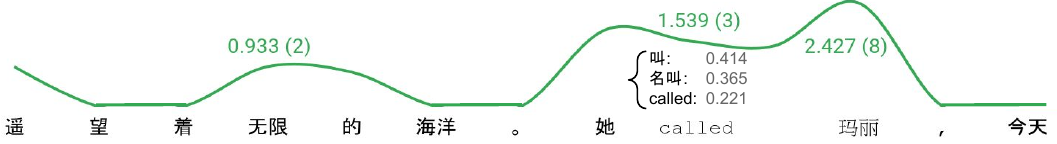}
     \caption{\textbf{A model is vulnerable to world-level language confusion when the number of tokens in the sampling nucleus is high, and the distribution is flat.} Metrics: Shannon entropy; in brackets: \# of tokens in nucleus.}
     \label{fig:sampling-peak}
\end{figure*}

\subsection{Impact of prompt length} 
We analyze whether the difficulty of our Complex prompts is caused by their much higher length (see Table \ref{tab:dataset-stats}) 
by grouping the prompts into three length buckets: short, medium and long, each with one third of the prompts. Table~\ref{tab:impact-prompt-length} shows the LPR of several models over the different lengths. We find no clear pattern, suggesting that higher confusion is caused by prompt complexity rather than length.

\subsection{Impact of instruction position}
Cross-lingual prompts include an instruction with the desired output language at the beginning, at the end, or integrated in the prompt (e.g., ``Write an essay in Korean [...]'').
Table~\ref{tab:impact-instruction-position} shows that across all models, line-level confusion is low for isolated instructions, with similar performance whether they are at the start or the end. Integrated instructions cause more confusion: Command R has only 69\% LPR on this type of prompt, versus $\sim$85\% for isolated types. This difficulty can be greatly reduced with one-shot prompting (80.6\% LPR; see \S\ref{sec:mitigation-instruction-tuning}).\footnote{Even though our demonstrations in few-shot prompting do not have integrated instructions.}

\subsection{Impact of quantization}
\label{sec:quant}
Quantization maps higher-precision LLM weights and activations to lower precision, reducing storage and inference costs, potentially at the cost of performance. We compare \textit{FP16} with \textit{W8}, \textit{W8A8}, and \textit{W4} variants of Command R+ on monolingual generation. Negative effects appear at \textit{W4}, shown in Table~\ref{tab:quant-short}. Details of quantization are in \S\ref{sec:app-quant}.

\subsection{Impact of instruction tuning} \label{sec:impact-instruction-tuning}

We compare the base with instruction-tuned variants of Llama and Command models on monolingual generation in Table \ref{tab:impact-instruction-tuning-monolingual-summary}. While instruction-tuned Command R models exhibit less language confusion than their base versions, instruction-tuned Llama models are much more confused, indicating English-centric instruction tuning, which is confirmed by our mitigation experiments (see \S\ref{sec:mitigation-instruction-tuning}).
\begin{table}[h]
\centering
\resizebox{\linewidth}{!}{%
\begin{tabular}{@{}larrrrrr@{}}
\toprule
                & avg & \multicolumn{1}{c}{ar} & \multicolumn{1}{c}{hi} & \multicolumn{1}{c}{ja} & \multicolumn{1}{c}{ko} & \multicolumn{1}{c}{vi} & \multicolumn{1}{c}{zh} \\ \midrule
Llama 2 70B     & \textbf{98.5}           & \textbf{99.6}          & \textbf{100.0}         & \textbf{100.0}         & \textbf{100.0}         & \textbf{98.0}          & \textbf{93.2}          \\
Llama 2 70B-I   & 6.0                     & 0.3                    & 1.0                    & 7.0                    & 0.0                    & 17.0                   & 10.5                   \\ \midrule
Llama 3 70B     & \textbf{94.7}           & \textbf{96.7}          & \textbf{97.9}          & \textbf{87.9}          & \textbf{98.8}          & \textbf{97.0}          & \textbf{90.0}          \\
Llama 3 70B-I   & 12.1                    & 21.7                   & 23.0                   & 10.0                   & 0.0                    & 10.0                   & 8.0                    \\ \midrule\midrule
Command R base  & 85.9                    & 94.9                   & 81.0                   & 93.9                   & 94.2                   & 83.0                   & 68.1                   \\
Command R       & \textbf{99.6}           & \textbf{100.0}         & \textbf{100.0}         & \textbf{100.0}         & \textbf{100.0}         & \textbf{99.0}          & \textbf{98.5}          \\ \midrule
Command R+ base & 78.4                    & 92.8                   & 67.0                   & 90.5                   & 93.5                   & 65.7                   & 60.9                   \\
Command R+      & \textbf{99.2}           & \textbf{99.7}          & \textbf{100.0}         & \textbf{99.0}          & \textbf{100.0}         & \textbf{99.0}          & \textbf{97.5}          \\ \bottomrule
\end{tabular}
}
\caption{\textbf{Line-level pass rate (LPR) of base vs instruction-tuned LLMs on monolingual generation} for a subset of languages. Full results in Table \ref{tab:impact-instruction-tuning-monolingual}, \S\ref{sec:base_vs_instruct}.}
\label{tab:impact-instruction-tuning-monolingual-summary}
\end{table}

\section{When does language confusion occur?} \label{sec:when-does-language-confusion-occur}

We study a sample of prompts to better understand where language confusion occurs. Intuitively, if a token in an undesired language is assigned sufficient probability, it may be sampled. We observe that language confusion typically occurs when the distribution over next tokens is flat and the nucleus is large (see \S\ref{sec:app-nuc-sampling} for background). 

We generate responses to 15 Chinese prompts from Okapi with Command R.\footnote{$k=0, p=0.75, t=0.7$} 
We examine outputs to identify instances of English language confusion, finding it in 5 of 15 outputs. 
For each, we find the first position where an English token was elicited, which we call the \textit{confusion point} (\textbf{CP}).\footnote{When intentional, such points are referred to in the code-switching literature as \textit{``switch points''}. 
We coin a new term here to indicate that this switching is erroneous.} 
There are 9 such CPs.\footnote{There are 3 examples of line-level confusion. We label the initial switching point from Chinese to English as the CP.}

We calculate Shannon entropy \citep{shannon1948mathematical} and nucleus size at each sampling point. We show an example output of Command R in Figure \ref{fig:sampling-peak}, indicating Shannon entropy and the number of tokens in the nucleus at select sampling points, and the next possible tokens with normalized probabilities at the confusion point. In the example, ``called'' was third most likely but occurred with sufficient probability to be sampled (0.221).
We generate 100 tokens per prompt, so there are 1500 points: 9 of which are CPs.  We refer to the others as \textbf{$\neg$CPs}.

\begin{figure*}[t]
    \centering
    \includegraphics[width=1\linewidth]{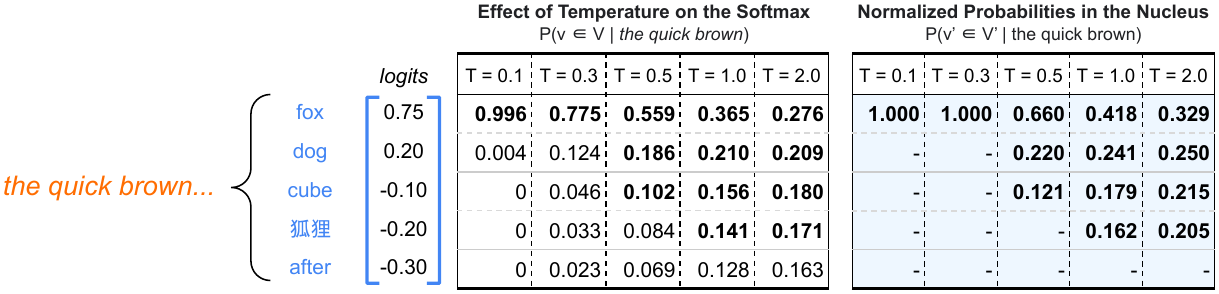}
     \caption{\textbf{Effect of Temperature ($T$) in Nucleus Sampling}. Tokens in the nucleus at $p=0.75$ are \textbf{bold}. Middle: Effect of $T$ on the softmax probabilities (Equation \ref{eq:softmax}). Right: Effect of $T$ on the probabilities of tokens in the nucleus right before sampling (Equation \ref{eq:normed_nucleus}). As $T$ increases, the token \begin{CJK*}{UTF8}{gbsn}狐狸\end{CJK*} has less chance to be sampled.}
     \label{fig:nucleus-sampling}
\end{figure*}

We average nucleus size and entropy over examples containing no instances of language confusion (10 samples, 1000 $\neg$CPs), at least one instance (5 examples, 9 CPs, 491 $\neg$CPs), and overall (15 samples, 9 CPs, 1491 $\neg$CPs). Results are in Table \ref{tab:lang-conf-point-avg-nuc-size}.\footnote{E.g., For outputs exhibiting language confusion, the average nucleus size @CP is: $\frac{\text{sum of nucleus sizes at all CPs}}{9} = \frac{32}{9} = 3.56$.} Outputs with and without language confusion show similar average nucleus size and entropy 
(1.64 vs. 1.61, and 0.353 vs. 0.365).  At CPs, however, average nucleus size and entropy is considerably higher: 3.56 nucleus size and 1.228 entropy vs 1.61/0.356 at $\neg$CPs, indicating that confusion tends to occur when nucleus size and entropy are high.

\begin{table}[htb]
\centering
\resizebox{\columnwidth}{!}{%
\begin{tabular}{@{}l|ccc|ccc@{}}
\toprule
        & \multicolumn{3}{c|}{\textbf{Avg. Nucleus Size}} & \multicolumn{3}{c}{\textbf{Avg. Entropy}} \\
 & Overall    & @CP    & $\neg$@CP   & Overall & @CP   & $\neg$@CP \\ \midrule
Has CP & 1.64       & 3.56   & 1.61                     & 0.353   & 1.228 & 0.337                  \\
No CP& 1.61       & -    & 1.61                     & 0.365 & -   & 0.365                  \\ \midrule
All     & 1.62       & 3.56   & 1.61                     & 0.361   & 1.228 & 0.356                  \\ \bottomrule
\end{tabular}%
}
\caption{\textbf{Avg. nucleus size, entropy at confusion points} (sampling points where language switch did [@CP] or did not [$\neg$@CP] occur) for 15 Chinese responses. Responses are split into those which had at least one CP (``Has CP'') or zero CPs (``No CP''). }
\label{tab:lang-conf-point-avg-nuc-size}
\end{table}

\section{Mitigating Language Confusion}
\label{sec:mitigating}
Based on the insights from our analyses in \S\ref{sec:impact-instruction-tuning} and \S\ref{sec:when-does-language-confusion-occur}, we propose \textit{inference-time} and \textit{training-time} mitigations for language confusion. 

\subsection{Reducing temperature and nucleus size}

Modifying sampling hyper-parameters affects which tokens are chosen at inference time. Section \ref{sec:app-nuc-sampling} explains nucleus sampling with temperature, and Figure \ref{fig:nucleus-sampling} has a toy example of manipulating the hyperparameter $T$. 
Framing language confusion as an undesired side-effect of sampling, it is intuitive that we might control it by sharpening the distribution over the next tokens at each timestep. 

We try to reduce the chance of having a wrong-language token sampled by manipulating temperature and nucleus size. Table \ref{tab:vary-pt} shows the results on monolingual WPR for Command R, with cross-lingual and LPR results in Section \ref{sec:app-nuc-samp-beam-search-extended}. Higher $T$ encourages language confusion: $T=1$ shows an average WPR of only 83.5\%, and as low as 72.0\% and 69.5\% for Japanese and Chinese.  Increasing $p$, resulting in a smaller nucleus, has a smaller effect. Note that setting $T=0.0$ is equivalent to sampling with $\text{top-K}=1$ (greedy search). 

\begin{table}[h]
\centering
\resizebox{\columnwidth}{!}{%
\begin{tabular}{@{}l|a|rrrrrr@{}}
\toprule
 & \multicolumn{1}{c}{avg} & \multicolumn{1}{c}{ar} & \multicolumn{1}{c}{hi} & \multicolumn{1}{c}{ja} & \multicolumn{1}{c}{ko} & \multicolumn{1}{c}{ru} & \multicolumn{1}{c}{zh} \\ \midrule
T=0.0 & \textbf{97.2} & 97.6 & \textbf{100.0} & \textbf{96.9} & 97.0 & \textbf{96.0} & \textbf{95.9} \\
T=0.3 & 96.3 & \textbf{99.3} & 99.0  & 93.9 & 97.0 & \textbf{96.0} & 92.3 \\
T=0.5 & 96.4 & 97.9 & 99.0  & 94.9 & \textbf{99.0} & 95.0 & 92.3 \\
T=0.7 & 94.2 & 98.0 & 98.0  & 91.8 & 93.9 & 93.0 & 90.3 \\
T=1.0 & \textcolor{red}{86.5} & \textcolor{red}{95.9} & \textcolor{red}{93.8}  & \textcolor{red}{74.5} & \textcolor{red}{92.8} & \textcolor{red}{87.5} & \textcolor{red}{74.7} \\ 
\midrule
p=0.1  & 97.4 & 98.3 & \textbf{100.0} & 98.0 & \textbf{97.0} & 95.0 & 95.9 \\
p=0.3  & 97.3 & \textcolor{red}{98.0} & \textbf{100.0} & \textbf{99.0} & \textbf{97.0} & \textcolor{red}{94.0} & 95.9 \\
p=0.5  & \textbf{97.6} & \textcolor{red}{98.0} & \textbf{100.0} & 96.9 & \textcolor{red}{96.0} & \textbf{98.0} & \textbf{96.9} \\
p=0.75 & \textcolor{red}{96.3} & \textbf{99.3} & 99.0  & \textcolor{red}{93.9} & \textbf{97.0} & 96.0 & \textcolor{red}{92.3} \\ \bottomrule 
\end{tabular}
}
\caption{\textbf{Effect of varying temperature ($T$) or nucleus size ($p$) on monolingual word-level language confusion (WPR) of \textit{Command R}}. Default values are $p=0.75$ and $T=0.3$. \textbf{Best score}. \textcolor{red}{Worst score.}}
\label{tab:vary-pt}
\end{table}




\subsection{Beam search decoding}
We explore the effect of beam search decoding on language confusion for Command R.  Table \ref{tab:beam-search} shows aggregate results for beam sizes 1 (greedy search) to 10, with full results in Section \ref{sec:app-nuc-samp-beam-search-extended}. Increasing beam size helps moderately over greedy search on WPR, with little noticable effect on monolingual LPR.
Increasing beam size consistently \textit{hurts} cross-lingual LPR and is most pronounced for non-Indo-European languages, where average LPR drops from 73.9 with greedy search to 65.6 with beam size = 10. 

\begin{table}[htb]
\centering
\resizebox{\linewidth}{!}{%
\begin{tabular}{@{}c|c@{\hspace{1.5ex}}c||r|r@{\hspace{1ex}}l|r@{\hspace{1ex}}l|r@{\hspace{1ex}}l@{}}
\toprule
& \multicolumn{2}{c||}{\textbf{Monoling.}} & \multicolumn{7}{c}{\textbf{Crosslingual}} \\
 & {WPR} & {LPR} & {WPR} & \multicolumn{6}{c}{LPR} \\ \midrule
 &  &  &  & \multicolumn{2}{c}{\textit{Overall}} & \multicolumn{2}{c}{\textit{$\neg$ IE}} & \multicolumn{2}{c}{\textit{IE}} \\
1 & 97.8 & \textbf{99.0} & 94.9 & \textbf{74.1} & - & \textbf{73.9} & - & \textbf{74.2} & - \\
2 & 98.6 & \textbf{99.0} & 95.4 & 72.2 & {\color[HTML]{EA4335} \textit{(-1.9)}} & 71.1 & {\color[HTML]{EA4335} \textit{(-2.8)}} & 73.6 & {\color[HTML]{EA4335} \textit{(-0.6)}} \\
3 & 98.6 & 98.7 & \textbf{97.1} & 71.5 & {\color[HTML]{EA4335} \textit{(-2.5)}} & 70.1 & {\color[HTML]{EA4335} \textit{(-3.8)}} & 73.4 & {\color[HTML]{EA4335} \textit{(-0.9)}} \\
5 & \textbf{99.0} & \textbf{99.0} & 96.7 & 70.3 & {\color[HTML]{EA4335} \textit{(-3.8)}} & 68.3 & {\color[HTML]{EA4335} \textit{(-5.6)}} & 72.9 & {\color[HTML]{EA4335} \textit{(-1.4)}} \\
10 & \textbf{99.0} & 98.5 & 96.7 & 68.4 & {\color[HTML]{EA4335} \textit{(-5.7)}} & 65.6 & {\color[HTML]{EA4335} \textit{(-8.3)}} & 72.1 & {\color[HTML]{EA4335} \textit{(-2.1)}} \\
\bottomrule
\end{tabular}%
}
\caption{\textbf{Effect of beam search decoding on language confusion metrics for \textit{Command R}.} Beam sizes: 1-10.}
\label{tab:beam-search}
\end{table}

Table \ref{tab:beam-v-nuc-samp} aggregates the best average score achieved for beam search and nucleus sampling from Tables \ref{tab:vary-pt}, \ref{tab:app-vary-pt-cross}, \ref{tab:beam-search-full-wpr}, \ref{tab:app-vary-pt-lpr}, \ref{tab:beam-search-full-lpr}.  Beam search is always better than nucleus sampling for WPR and for cross-lingual LPR, and both methods seem equally effective for monolingual LPR, suggesting that beam search may be an effective decoding strategy for lessening language confusion (though at higher computational cost).

\begin{table}[htb]
\centering
\resizebox{\columnwidth}{!}{%
\begin{tabular}{@{}l|cc|cc||cc|cc@{}}
\toprule
 & \multicolumn{4}{c||}{\textbf{WPR}} & \multicolumn{4}{c}{\textbf{LPR}} \\
 & \multicolumn{2}{c}{\textbf{Beam S.}} & \multicolumn{2}{c||}{\textbf{Nuc. Samp.}} & \multicolumn{2}{c}{\textbf{Beam S.}} & \multicolumn{2}{c}{\textbf{Nuc. Samp.}} \\
 & \textit{max} & \textit{min} & \textit{max} & \textit{min} & \textit{max} & \textit{min} & \textit{max} & \textit{min} \\ \midrule
Mono. & \textbf{99.0} & \textit{97.8} & 97.6 & 86.5 & \textbf{99.0} & 98.5 & \textit{98.9} & 96.8 \\ \midrule
Cross. & \textbf{97.1} & \textit{94.9} & 94.6 & 80.1 & \textbf{74.1} & \textit{68.4} & 68.2 & 64.0 \\ \bottomrule
\end{tabular}%
}
\caption{Beam Search vs. Nucleus Sampling. Shown: \textbf{best} and \textit{second-best} overall average score.}
\label{tab:beam-v-nuc-samp}
\end{table}

\subsection{Few-shot prompting}

Non-instruction-tuned LLMs often do not answer instructions as expected. When prompted with an instruction in a non-English language, for instance, Command R Base often translates it instead of answering. We thus employ few-shot prompting to guide Command R Base towards the correct behavior. We cherry-pick 5 prompt/answer pairs in English and translate them with Google Translate.\footnote{Few-shot demonstrations are in different languages than the current task's target language. The goal is to help a model understand the task, not find the language to generate into.} 

For cross-lingual generation, we use a similar format to the test sets. Figure~\ref{fig:few_shot_template} shows the template used to prompt the base model. For comparison, we also apply one-shot prompting to Command R where the example is presented as turns in a conversation.
Table~\ref{tab:ablation} shows the results. Tables~\ref{tab:ablation_monolingual_detailed} and~\ref{tab:ablation_crosslingual_detailed} in Appendix show the detailed results per language for monolingual and cross-lingual language confusion respectively.

Few-shot prompting greatly reduces Command R Base's language confusion and almost completely eliminates the problem in the monolingual setting. While one-shot prompting with Command R is detrimental monolingually---indicating a difficulty of dealing with demonstrations in other languages---it enables the model to better follow the cross-lingual instructions.

\begin{table}[h]
\centering
\resizebox{\linewidth}{!}{%
\begin{tabular}{lrr|rr}
\toprule
 & \multicolumn{2}{c|}{Monolingual} & \multicolumn{2}{c}{Cross-lingual} \\ \midrule
 & LPR & WPR & LPR & WPR \\ \midrule
Command R Base & 86.2 & 98.7 & 1.1 & \textbf{100.0} \\
+ Q/A template (0-shot) & 85.3 & 99.7 & 20.9 & 97.0 \\
+ 1-shot & 94.1 & \textbf{100.0} & 90.7 & 98.6 \\
+ 5-shot & \textbf{99.0} & \textbf{100.0} & \textbf{95.0} & 99.7 \\
+ English SFT & 77.8 & 96.2 & 78.3 & 91.7 \\
\quad{} + English pref. tuning & 74.3 & 90.9 & 85.7 & 87.4 \\
+ Multilingual SFT & 98.3 & 95.5 & 78.2 & 90.0 \\
\quad{} + Multi. pref. tuning & 98.9 & 93.4 & 89.4 & 86.9 \\
\hline
\emph{Command R} & 98.6 & 96.3 & 68.1 & 94.0 \\
+ 1-shot & 68.3 & 92.7 & 82.9 & 92.3 \\
\bottomrule
\end{tabular}%
}
\caption{\textbf{Effect of few-shot prompting and instruction tuning on language confusion.}}
\label{tab:ablation}
\end{table}

\subsection{Multilingual instruction tuning} \label{sec:mitigation-instruction-tuning}
We additionally study the impact of English-centric and multilingual instruction tuning by fine-tuning Command R Base with several techniques:
\paragraph{English-only tuning} We fine-tune the base model with English-only publicly available instruction data \cite[SFT;][]{touvron2023llama}, then apply English-only preference tuning to this model.
\paragraph{Multilingual tuning} We extend the English data with multilingual data, most of which comes from machine-translated Dolly and ShareGPT \cite{ustun2024aya}. Due to the scarcity of multilingual data, our SFT data mixture is 90\% English. For preference tuning, we use 50\% multilingual data. Results are in Table~\ref{tab:ablation}.


English-only tuning (both SFT and pref. tuning) exacerbates language confusion monolingually: likely the reason for high language confusion of Llama-Instruct models (see \S\ref{sec:impact-instruction-tuning}). English-only preference tuning has a strong negative impact on word-level confusion. However, SFT with just 10\% multilingual data is enough to almost eliminate the problem of line-level confusion monolingually. Cross-lingually, multilingual tuning does not give better line-level performance than English-only tuning. This may be because cross-lingual datasets only have English prompts and it is more important for the model to learn to follow English instructions (e.g., ``Reply in French'').



\section{Discussion}

We discuss several aspects that raise important questions including the behavior of base models, the English centricity of models, the impact of preference tuning, and the importance of other factors in Appendix \ref{app:discussion}.

\section{Related Work}

\paragraph{Code-switching} There has been much research on \emph{natural} alternations between languages, i.e., code-switching \cite{dogruoz-etal-2021-survey} in natural language processing (NLP). Prior work focused on evaluating capabilities on standard NLP tasks using code-switching data \cite{khanuja-etal-2020-gluecos,winata-etal-2023-decades} including sentiment analysis, machine translation, summarization and word-level language identification. These tasks typically employ data created by humans where word-level code-switching occurs between English and another language such as Hindi, Spanish, or Arabic. Current models still struggle to generate and understand code-switched text in some languages \cite{yong-etal-2023-prompting,zhang-etal-2023-multilingual}. In contrast, we investigate \emph{unnatural} and \emph{erroneous} code-switching---or language confusion---in the LLM's generations.

\paragraph{Language confusion} Prior work has observed `source language hallucinations' in zero-shot cross-lingual transfer \cite{vu-etal-2022-overcoming,li-murray-2023-zero,pfeiffer-etal-2023-mmt5,chirkova-nikoulina-2024-key} when models are fine-tuned on English data and applied to generate text in another language.
The problem of language confusion is known in the machine translation field as `off-target translation' \cite{chen-etal-2023-target,sennrich-etal-2024-mitigating}. It typically occurs on English-centric multilingual models, when used in a zero-shot manner (to translate between two languages unseen at training).
For LLMs, there is no source language per se. A few studies \cite{kew2023turning,faisal-anastasopoulos-2023-geographic,chen-etal-2024-monolingual} have provided evidence of LLMs generating in an incorrect language on the response level. \citet{holtermann2024evaluating} analyze which languages are confused on the response level using mainly smaller LLMs. To the best of our knowledge, we are the first to show results on the line and word level and the first to systematically study language confusion in LLMs.


\section{Conclusion}

We have introduced the Language Confusion Benchmark. We have shown that some LLMs exhibit severe language confusion and even the strongest LLMs do not achieve perfect performance cross-lingually. We observed that base and English-centric instruct models are particularly susceptible to language confusion, which is exacerbated by complex prompts. Finally, we proposed measures to mitigate language confusion at inference and training. Our benchmark is efficient to evaluate and easy to extend and can help ensure that models achieve equal utility across languages.

\section*{Acknowledgements}

We thank Sander Land, Florian Strub, Mo Gheshlaghi Azar, and Olivier Pietquin for early discussions on language confusion. We are grateful to Sander Land for help with an initial version of a script to analyze nucleus sampling generations. We thank our internal annotators and the team at Cohere.

\section*{Limitations}

As this is to our knowledge, the first systematic study on language confusion in LLMs, there are several natural research directions that we were not able to cover:

\paragraph{Conversations} We focus on single-turn inputs, and do not consider the impact of multiple turns of a conversation or users employing different languages across different turns.

\paragraph{Code-switched input} We focus on the setting where a user's prompt is in a single language (the target language in the monolingual setting and English in the cross-lingual setting). We do not consider inputs that are naturally code-switched.

\paragraph{Inputs with cross-lingual context} Content may not always be available in a user's language, which is relevant for applications such as cross-lingual summarization or cross-lingual QA.

\paragraph{Language varieties} We evaluate generation into standardized languages. Future work may expand to language varieties and dialects, styles, and registers.

Our metrics are applied to model outputs of at most 100 tokens. But what is a token depends on the model's tokenizer. Models with more aggressive tokenization could be advantaged with regard to the binary metrics (LPR and WPR). Likewise, because we allow models to stop their generation early (by producing an end-of-sequence token), models which are less verbose could have an advantage.  Furthermore, because current LID tools do not support word-level language identification, our WPR metric is currently limited to assessing non-Latin script languages for unintended switching into English.  As LID tools 

\bibliography{main}

\begin{thebibliography}{49}
\providecommand{\natexlab}[1]{#1}

\bibitem[{Achiam et~al.(2023)Achiam, Adler, Agarwal, Ahmad, Akkaya, Aleman, Almeida, Altenschmidt, Altman, Anadkat et~al.}]{achiam2023gpt}
Josh Achiam, Steven Adler, Sandhini Agarwal, Lama Ahmad, Ilge Akkaya, Florencia~Leoni Aleman, Diogo Almeida, Janko Altenschmidt, Sam Altman, Shyamal Anadkat, et~al. 2023.
\newblock Gpt-4 technical report.
\newblock \emph{arXiv preprint arXiv:2303.08774}.

\bibitem[{Ahia et~al.(2023)Ahia, Kumar, Gonen, Kasai, Mortensen, Smith, and Tsvetkov}]{ahia-etal-2023-languages}
Orevaoghene Ahia, Sachin Kumar, Hila Gonen, Jungo Kasai, David Mortensen, Noah Smith, and Yulia Tsvetkov. 2023.
\newblock \href {https://doi.org/10.18653/v1/2023.emnlp-main.614} {Do all languages cost the same? tokenization in the era of commercial language models}.
\newblock In \emph{Proceedings of the 2023 Conference on Empirical Methods in Natural Language Processing}, pages 9904--9923, Singapore. Association for Computational Linguistics.

\bibitem[{Asai et~al.(2023)Asai, Kudugunta, Yu, Blevins, Gonen, Reid, Tsvetkov, Ruder, and Hajishirzi}]{asai2023buffet}
Akari Asai, Sneha Kudugunta, Xinyan~Velocity Yu, Terra Blevins, Hila Gonen, Machel Reid, Yulia Tsvetkov, Sebastian Ruder, and Hannaneh Hajishirzi. 2023.
\newblock Buffet: Benchmarking large language models for few-shot cross-lingual transfer.
\newblock \emph{arXiv preprint arXiv:2305.14857}.

\bibitem[{Bang et~al.(2023)Bang, Cahyawijaya, Lee, Dai, Su, Wilie, Lovenia, Ji, Yu, Chung, Do, Xu, and Fung}]{bang-etal-2023-multitask}
Yejin Bang, Samuel Cahyawijaya, Nayeon Lee, Wenliang Dai, Dan Su, Bryan Wilie, Holy Lovenia, Ziwei Ji, Tiezheng Yu, Willy Chung, Quyet~V. Do, Yan Xu, and Pascale Fung. 2023.
\newblock \href {https://doi.org/10.18653/v1/2023.ijcnlp-main.45} {A multitask, multilingual, multimodal evaluation of {C}hat{GPT} on reasoning, hallucination, and interactivity}.
\newblock In \emph{Proceedings of the 13th International Joint Conference on Natural Language Processing and the 3rd Conference of the Asia-Pacific Chapter of the Association for Computational Linguistics (Volume 1: Long Papers)}, pages 675--718, Nusa Dua, Bali. Association for Computational Linguistics.

\bibitem[{Blevins and Zettlemoyer(2022)}]{blevins-zettlemoyer-2022-language}
Terra Blevins and Luke Zettlemoyer. 2022.
\newblock \href {https://doi.org/10.18653/v1/2022.emnlp-main.233} {Language contamination helps explains the cross-lingual capabilities of {E}nglish pretrained models}.
\newblock In \emph{Proceedings of the 2022 Conference on Empirical Methods in Natural Language Processing}, pages 3563--3574, Abu Dhabi, United Arab Emirates. Association for Computational Linguistics.

\bibitem[{Brown et~al.(2020)Brown, Mann, Ryder, Subbiah, Kaplan, Dhariwal, Neelakantan, Shyam, Sastry, Askell et~al.}]{brown2020language}
Tom Brown, Benjamin Mann, Nick Ryder, Melanie Subbiah, Jared~D Kaplan, Prafulla Dhariwal, Arvind Neelakantan, Pranav Shyam, Girish Sastry, Amanda Askell, et~al. 2020.
\newblock Language models are few-shot learners.
\newblock \emph{Advances in neural information processing systems}, 33:1877--1901.

\bibitem[{Chen et~al.(2023)Chen, Ma, Zhang, Wei, and Chang}]{chen-etal-2023-target}
Liang Chen, Shuming Ma, Dongdong Zhang, Furu Wei, and Baobao Chang. 2023.
\newblock \href {https://doi.org/10.18653/v1/2023.findings-acl.608} {On the off-target problem of zero-shot multilingual neural machine translation}.
\newblock In \emph{Findings of the Association for Computational Linguistics: ACL 2023}, pages 9542--9558, Toronto, Canada. Association for Computational Linguistics.

\bibitem[{Chen et~al.(2024)Chen, Ji, Bogoychev, Kutuzov, Haddow, and Heafield}]{chen-etal-2024-monolingual}
Pinzhen Chen, Shaoxiong Ji, Nikolay Bogoychev, Andrey Kutuzov, Barry Haddow, and Kenneth Heafield. 2024.
\newblock \href {https://aclanthology.org/2024.findings-eacl.90} {Monolingual or multilingual instruction tuning: Which makes a better alpaca}.
\newblock In \emph{Findings of the Association for Computational Linguistics: EACL 2024}, pages 1347--1356, St. Julian{'}s, Malta. Association for Computational Linguistics.

\bibitem[{Chirkova and Nikoulina(2024)}]{chirkova-nikoulina-2024-key}
Nadezhda Chirkova and Vassilina Nikoulina. 2024.
\newblock \href {https://aclanthology.org/2024.naacl-long.401} {Key ingredients for effective zero-shot cross-lingual knowledge transfer in generative tasks}.
\newblock In \emph{Proceedings of the 2024 Conference of the North American Chapter of the Association for Computational Linguistics: Human Language Technologies (Volume 1: Long Papers)}, pages 7222--7238, Mexico City, Mexico. Association for Computational Linguistics.

\bibitem[{Conover et~al.(2023)Conover, Hayes, Mathur, Xie, Wan, Shah, Ghodsi, Wendell, Zaharia, and Xin}]{conover2023free}
Mike Conover, Matt Hayes, Ankit Mathur, Jianwei Xie, Jun Wan, Sam Shah, Ali Ghodsi, Patrick Wendell, Matei Zaharia, and Reynold Xin. 2023.
\newblock \href {https://www.databricks.com/blog/2023/04/12/dolly-first-open-commercially-viable-instruction-tuned-llm} {Free dolly: Introducing the world’s first truly open instruction-tuned llm}.
\newblock \emph{Company Blog of Databricks}.

\bibitem[{Do{\u{g}}ru{\"o}z et~al.(2021)Do{\u{g}}ru{\"o}z, Sitaram, Bullock, and Toribio}]{dogruoz-etal-2021-survey}
A.~Seza Do{\u{g}}ru{\"o}z, Sunayana Sitaram, Barbara~E. Bullock, and Almeida~Jacqueline Toribio. 2021.
\newblock \href {https://doi.org/10.18653/v1/2021.acl-long.131} {A survey of code-switching: Linguistic and social perspectives for language technologies}.
\newblock In \emph{Proceedings of the 59th Annual Meeting of the Association for Computational Linguistics and the 11th International Joint Conference on Natural Language Processing (Volume 1: Long Papers)}, pages 1654--1666, Online. Association for Computational Linguistics.

\bibitem[{Dubey et~al.(2024)Dubey, Jauhri, Pandey, Kadian, Al-Dahle, Letman, Mathur, Schelten, Yang, Fan et~al.}]{dubey2024llama}
Abhimanyu Dubey, Abhinav Jauhri, Abhinav Pandey, Abhishek Kadian, Ahmad Al-Dahle, Aiesha Letman, Akhil Mathur, Alan Schelten, Amy Yang, Angela Fan, et~al. 2024.
\newblock The llama 3 herd of models.
\newblock \emph{arXiv preprint arXiv:2407.21783}.

\bibitem[{Faisal and Anastasopoulos(2023)}]{faisal-anastasopoulos-2023-geographic}
Fahim Faisal and Antonios Anastasopoulos. 2023.
\newblock \href {https://doi.org/10.18653/v1/2023.mrl-1.12} {Geographic and geopolitical biases of language models}.
\newblock In \emph{Proceedings of the 3rd Workshop on Multi-lingual Representation Learning (MRL)}, pages 139--163, Singapore. Association for Computational Linguistics.

\bibitem[{Frantar et~al.(2022)Frantar, Ashkboos, Hoefler, and Alistarh}]{frantar-gptq}
Elias Frantar, Saleh Ashkboos, Torsten Hoefler, and Dan Alistarh. 2022.
\newblock {GPTQ}: Accurate post-training compression for generative pretrained transformers.
\newblock \emph{arXiv preprint arXiv:2210.17323}.

\bibitem[{Grice(1975)}]{grice1975logic}
Herbert~P Grice. 1975.
\newblock Logic and conversation.
\newblock In \emph{Speech acts}, pages 41--58. Brill.

\bibitem[{Gudibande et~al.(2024)Gudibande, Wallace, Snell, Geng, Liu, Abbeel, Levine, and Song}]{gudibande2024false}
Arnav Gudibande, Eric Wallace, Charlie~Victor Snell, Xinyang Geng, Hao Liu, Pieter Abbeel, Sergey Levine, and Dawn Song. 2024.
\newblock The false promise of imitating proprietary language models.
\newblock In \emph{Proceedings of ICLR 2024}.

\bibitem[{Guerreiro et~al.(2023)Guerreiro, Alves, Waldendorf, Haddow, Birch, Colombo, and Martins}]{guerreiro2023hallucinations}
Nuno~M Guerreiro, Duarte~M Alves, Jonas Waldendorf, Barry Haddow, Alexandra Birch, Pierre Colombo, and Andr{\'e}~FT Martins. 2023.
\newblock Hallucinations in large multilingual translation models.
\newblock \emph{Transactions of the Association for Computational Linguistics}, 11:1500--1517.

\bibitem[{Held et~al.(2023)Held, Harris, Best, and Yang}]{held2023material}
William Held, Camille Harris, Michael Best, and Diyi Yang. 2023.
\newblock A material lens on coloniality in nlp.
\newblock \emph{arXiv preprint arXiv:2311.08391}.

\bibitem[{Holtermann et~al.(2024)Holtermann, R{\"o}ttger, Dill, and Lauscher}]{holtermann2024evaluating}
Carolin Holtermann, Paul R{\"o}ttger, Timm Dill, and Anne Lauscher. 2024.
\newblock Evaluating the elementary multilingual capabilities of large language models with multiq.
\newblock \emph{arXiv preprint arXiv:2403.03814}.

\bibitem[{Holtzman et~al.(2019)Holtzman, Buys, Du, Forbes, and Choi}]{holtzman2019curious}
Ari Holtzman, Jan Buys, Li~Du, Maxwell Forbes, and Yejin Choi. 2019.
\newblock The curious case of neural text degeneration.
\newblock In \emph{International Conference on Learning Representations}.

\bibitem[{Hosking et~al.(2024)Hosking, Blunsom, and Bartolo}]{hosking2024human}
Tom Hosking, Phil Blunsom, and Max Bartolo. 2024.
\newblock Human feedback is not gold standard.
\newblock In \emph{Proceedings of ICLR 2024}.

\bibitem[{Hu et~al.(2020)Hu, Ruder, Siddhant, Neubig, Firat, and Johnson}]{hu2020xtreme}
Junjie Hu, Sebastian Ruder, Aditya Siddhant, Graham Neubig, Orhan Firat, and Melvin Johnson. 2020.
\newblock Xtreme: A massively multilingual multi-task benchmark for evaluating cross-lingual generalisation.
\newblock In \emph{International Conference on Machine Learning}, pages 4411--4421. PMLR.

\bibitem[{Ji et~al.(2023)Ji, Yu, Xu, Lee, Ishii, and Fung}]{ji2023towards}
Ziwei Ji, Tiezheng Yu, Yan Xu, Nayeon Lee, Etsuko Ishii, and Pascale Fung. 2023.
\newblock Towards mitigating llm hallucination via self reflection.
\newblock In \emph{Findings of the Association for Computational Linguistics: EMNLP 2023}, pages 1827--1843.

\bibitem[{Jiang et~al.(2024)Jiang, Sablayrolles, Roux, Mensch, Savary, Bamford, Chaplot, Casas, Hanna, Bressand et~al.}]{jiang2024mixtral}
Albert~Q Jiang, Alexandre Sablayrolles, Antoine Roux, Arthur Mensch, Blanche Savary, Chris Bamford, Devendra~Singh Chaplot, Diego de~las Casas, Emma~Bou Hanna, Florian Bressand, et~al. 2024.
\newblock Mixtral of experts.
\newblock \emph{arXiv preprint arXiv:2401.04088}.

\bibitem[{Joshi et~al.(2020)Joshi, Santy, Budhiraja, Bali, and Choudhury}]{joshi-etal-2020-state}
Pratik Joshi, Sebastin Santy, Amar Budhiraja, Kalika Bali, and Monojit Choudhury. 2020.
\newblock \href {https://doi.org/10.18653/v1/2020.acl-main.560} {The state and fate of linguistic diversity and inclusion in the {NLP} world}.
\newblock In \emph{Proceedings of the 58th Annual Meeting of the Association for Computational Linguistics}, pages 6282--6293, Online. Association for Computational Linguistics.

\bibitem[{Joulin et~al.(2016)Joulin, Grave, Bojanowski, and Mikolov}]{joulin2016bag}
Armand Joulin, Edouard Grave, Piotr Bojanowski, and Tomas Mikolov. 2016.
\newblock Bag of tricks for efficient text classification.
\newblock \emph{arXiv preprint arXiv:1607.01759}.

\bibitem[{Kaddour et~al.(2023)Kaddour, Harris, Mozes, Bradley, Raileanu, and McHardy}]{kaddour2023challenges}
Jean Kaddour, Joshua Harris, Maximilian Mozes, Herbie Bradley, Roberta Raileanu, and Robert McHardy. 2023.
\newblock Challenges and applications of large language models.
\newblock \emph{arXiv preprint arXiv:2307.10169}.

\bibitem[{Kantar and IAMAI(2023)}]{kantar_internet_2023}
Kantar and IAMAI. 2023.
\newblock \href {https://www.iamai.in/sites/default/files/research/INTERNET%20IN%20INDIA%202023.pdf} {Internet in {India} 2023}.
\newblock Technical report, Internet And Mobile Association of India.

\bibitem[{Kew et~al.(2023)Kew, Schottmann, and Sennrich}]{kew2023turning}
Tannon Kew, Florian Schottmann, and Rico Sennrich. 2023.
\newblock Turning english-centric llms into polyglots: How much multilinguality is needed?
\newblock \emph{arXiv preprint arXiv:2312.12683}.

\bibitem[{Khanuja et~al.(2020)Khanuja, Dandapat, Srinivasan, Sitaram, and Choudhury}]{khanuja-etal-2020-gluecos}
Simran Khanuja, Sandipan Dandapat, Anirudh Srinivasan, Sunayana Sitaram, and Monojit Choudhury. 2020.
\newblock \href {https://doi.org/10.18653/v1/2020.acl-main.329} {{GLUEC}o{S}: An evaluation benchmark for code-switched {NLP}}.
\newblock In \emph{Proceedings of the 58th Annual Meeting of the Association for Computational Linguistics}, pages 3575--3585, Online. Association for Computational Linguistics.

\bibitem[{Lai et~al.(2023)Lai, Nguyen, Ngo, Nguyen, Dernoncourt, Rossi, and Nguyen}]{lai-etal-2023-okapi}
Viet Lai, Chien Nguyen, Nghia Ngo, Thuat Nguyen, Franck Dernoncourt, Ryan Rossi, and Thien Nguyen. 2023.
\newblock \href {https://doi.org/10.18653/v1/2023.emnlp-demo.28} {Okapi: Instruction-tuned large language models in multiple languages with reinforcement learning from human feedback}.
\newblock In \emph{Proceedings of the 2023 Conference on Empirical Methods in Natural Language Processing: System Demonstrations}, pages 318--327, Singapore. Association for Computational Linguistics.

\bibitem[{Li and Murray(2023)}]{li-murray-2023-zero}
Tianjian Li and Kenton Murray. 2023.
\newblock \href {https://doi.org/10.18653/v1/2023.findings-acl.789} {Why does zero-shot cross-lingual generation fail? an explanation and a solution}.
\newblock In \emph{Findings of the Association for Computational Linguistics: ACL 2023}, pages 12461--12476, Toronto, Canada. Association for Computational Linguistics.

\bibitem[{Pfeiffer et~al.(2023)Pfeiffer, Piccinno, Nicosia, Wang, Reid, and Ruder}]{pfeiffer-etal-2023-mmt5}
Jonas Pfeiffer, Francesco Piccinno, Massimo Nicosia, Xinyi Wang, Machel Reid, and Sebastian Ruder. 2023.
\newblock \href {https://doi.org/10.18653/v1/2023.findings-emnlp.132} {mm{T}5: Modular multilingual pre-training solves source language hallucinations}.
\newblock In \emph{Findings of the Association for Computational Linguistics: EMNLP 2023}, pages 1978--2008, Singapore. Association for Computational Linguistics.

\bibitem[{Rafailov et~al.(2023)Rafailov, Sharma, Mitchell, Manning, Ermon, and Finn}]{rafailov_dpo}
Rafael Rafailov, Archit Sharma, Eric Mitchell, Christopher~D Manning, Stefano Ermon, and Chelsea Finn. 2023.
\newblock \href {https://proceedings.neurips.cc/paper_files/paper/2023/file/a85b405ed65c6477a4fe8302b5e06ce7-Paper-Conference.pdf} {Direct preference optimization: Your language model is secretly a reward model}.
\newblock In \emph{Advances in Neural Information Processing Systems}, volume~36, pages 53728--53741. Curran Associates, Inc.

\bibitem[{Sennrich et~al.(2024)Sennrich, Vamvas, and Mohammadshahi}]{sennrich-etal-2024-mitigating}
Rico Sennrich, Jannis Vamvas, and Alireza Mohammadshahi. 2024.
\newblock \href {https://aclanthology.org/2024.eacl-short.4} {Mitigating hallucinations and off-target machine translation with source-contrastive and language-contrastive decoding}.
\newblock In \emph{Proceedings of the 18th Conference of the European Chapter of the Association for Computational Linguistics (Volume 2: Short Papers)}, pages 21--33, St. Julian{'}s, Malta. Association for Computational Linguistics.

\bibitem[{Shannon(1948)}]{shannon1948mathematical}
Claude~Elwood Shannon. 1948.
\newblock A mathematical theory of communication.
\newblock \emph{The Bell system technical journal}, 27(3):379--423.

\bibitem[{Singh et~al.(2024)Singh, Vargus, Dsouza, Karlsson, Mahendiran, Ko, Shandilya, Patel, Mataciunas, OMahony et~al.}]{singh2024aya}
Shivalika Singh, Freddie Vargus, Daniel Dsouza, B{\"o}rje~F Karlsson, Abinaya Mahendiran, Wei-Yin Ko, Herumb Shandilya, Jay Patel, Deividas Mataciunas, Laura OMahony, et~al. 2024.
\newblock Aya dataset: An open-access collection for multilingual instruction tuning.
\newblock \emph{arXiv preprint arXiv:2402.06619}.

\bibitem[{Taori et~al.(2023)Taori, Gulrajani, Zhang, Dubois, Li, Guestrin, Liang, and Hashimoto}]{taori2023alpaca}
Rohan Taori, Ishaan Gulrajani, Tianyi Zhang, Yann Dubois, Xuechen Li, Carlos Guestrin, Percy Liang, and Tatsunori~B Hashimoto. 2023.
\newblock \href {https://crfm.stanford.edu/2023/03/13/alpaca.html} {Alpaca: A strong, replicable instruction-following model}.
\newblock \emph{Stanford Center for Research on Foundation Models}.

\bibitem[{Touvron et~al.(2023)Touvron, Martin, Stone, Albert, Almahairi, Babaei, Bashlykov, Batra, Bhargava, Bhosale et~al.}]{touvron2023llama}
Hugo Touvron, Louis Martin, Kevin Stone, Peter Albert, Amjad Almahairi, Yasmine Babaei, Nikolay Bashlykov, Soumya Batra, Prajjwal Bhargava, Shruti Bhosale, et~al. 2023.
\newblock Llama 2: Open foundation and fine-tuned chat models.
\newblock \emph{arXiv preprint arXiv:2307.09288}.

\bibitem[{{\"U}st{\"u}n et~al.(2024){\"U}st{\"u}n, Aryabumi, Yong, Ko, D'souza, Onilude, Bhandari, Singh, Ooi, Kayid et~al.}]{ustun2024aya}
Ahmet {\"U}st{\"u}n, Viraat Aryabumi, Zheng-Xin Yong, Wei-Yin Ko, Daniel D'souza, Gbemileke Onilude, Neel Bhandari, Shivalika Singh, Hui-Lee Ooi, Amr Kayid, et~al. 2024.
\newblock Aya model: An instruction finetuned open-access multilingual language model.
\newblock \emph{arXiv preprint arXiv:2402.07827}.

\bibitem[{Vu et~al.(2022)Vu, Barua, Lester, Cer, Iyyer, and Constant}]{vu-etal-2022-overcoming}
Tu~Vu, Aditya Barua, Brian Lester, Daniel Cer, Mohit Iyyer, and Noah Constant. 2022.
\newblock \href {https://doi.org/10.18653/v1/2022.emnlp-main.630} {Overcoming catastrophic forgetting in zero-shot cross-lingual generation}.
\newblock In \emph{Proceedings of the 2022 Conference on Empirical Methods in Natural Language Processing}, pages 9279--9300, Abu Dhabi, United Arab Emirates. Association for Computational Linguistics.

\bibitem[{Wilson and Sperber(2012)}]{wilson2012linguistic}
Deirdre Wilson and Dan Sperber. 2012.
\newblock Linguistic form and relevance.
\newblock \emph{Wilson \& Sperber (Eds.), Meaning and Relevance}, pages 149--168.

\bibitem[{Winata et~al.(2023)Winata, Aji, Yong, and Solorio}]{winata-etal-2023-decades}
Genta Winata, Alham~Fikri Aji, Zheng~Xin Yong, and Thamar Solorio. 2023.
\newblock \href {https://doi.org/10.18653/v1/2023.findings-acl.185} {The decades progress on code-switching research in {NLP}: A systematic survey on trends and challenges}.
\newblock In \emph{Findings of the Association for Computational Linguistics: ACL 2023}, pages 2936--2978, Toronto, Canada. Association for Computational Linguistics.

\bibitem[{Xu et~al.(2024)Xu, Fu, Gao, Ye, Liu, Mei, Wang, Yu, and Wu}]{xu2024dposuperiorppollm}
Shusheng Xu, Wei Fu, Jiaxuan Gao, Wenjie Ye, Weilin Liu, Zhiyu Mei, Guangju Wang, Chao Yu, and Yi~Wu. 2024.
\newblock \href {https://arxiv.org/abs/2404.10719} {Is dpo superior to ppo for llm alignment? a comprehensive study}.
\newblock \emph{Preprint}, arXiv:2404.10719.

\bibitem[{Yan et~al.(2024)Yan, Miao, Li, Zhang, Xie, Deng, and Yan}]{yan20243dpropertiesidentifyingchallengesdpo}
Yuzi Yan, Yibo Miao, Jialian Li, Yipin Zhang, Jian Xie, Zhijie Deng, and Dong Yan. 2024.
\newblock \href {https://arxiv.org/abs/2406.07327} {3d-properties: Identifying challenges in dpo and charting a path forward}.
\newblock \emph{Preprint}, arXiv:2406.07327.

\bibitem[{Yong et~al.(2023)Yong, Zhang, Forde, Wang, Subramonian, Lovenia, Cahyawijaya, Winata, Sutawika, Cruz, Tan, Phan, Phan, Garcia, Solorio, and Aji}]{yong-etal-2023-prompting}
Zheng~Xin Yong, Ruochen Zhang, Jessica Forde, Skyler Wang, Arjun Subramonian, Holy Lovenia, Samuel Cahyawijaya, Genta Winata, Lintang Sutawika, Jan Christian~Blaise Cruz, Yin~Lin Tan, Long Phan, Long Phan, Rowena Garcia, Thamar Solorio, and Alham Aji. 2023.
\newblock \href {https://doi.org/10.18653/v1/2023.calcs-1.5} {Prompting multilingual large language models to generate code-mixed texts: The case of south {E}ast {A}sian languages}.
\newblock In \emph{Proceedings of the 6th Workshop on Computational Approaches to Linguistic Code-Switching}, pages 43--63, Singapore. Association for Computational Linguistics.

\bibitem[{Yuan et~al.(2024)Yuan, Cui, Wang, Ding, Wang, Deng, Shan, Chen, Xie, Lin, Liu, Zhou, Peng, Liu, and Sun}]{yuan2024advancingllmreasoninggeneralists}
Lifan Yuan, Ganqu Cui, Hanbin Wang, Ning Ding, Xingyao Wang, Jia Deng, Boji Shan, Huimin Chen, Ruobing Xie, Yankai Lin, Zhenghao Liu, Bowen Zhou, Hao Peng, Zhiyuan Liu, and Maosong Sun. 2024.
\newblock \href {https://arxiv.org/abs/2404.02078} {Advancing llm reasoning generalists with preference trees}.
\newblock \emph{Preprint}, arXiv:2404.02078.

\bibitem[{Zhang et~al.(2023)Zhang, Cahyawijaya, Cruz, Winata, and Aji}]{zhang-etal-2023-multilingual}
Ruochen Zhang, Samuel Cahyawijaya, Jan Christian~Blaise Cruz, Genta Winata, and Alham Aji. 2023.
\newblock \href {https://doi.org/10.18653/v1/2023.emnlp-main.774} {Multilingual large language models are not (yet) code-switchers}.
\newblock In \emph{Proceedings of the 2023 Conference on Empirical Methods in Natural Language Processing}, pages 12567--12582, Singapore. Association for Computational Linguistics.

\bibitem[{Zhao et~al.(2024)Zhao, Zhang, Zhang, Gui, and Huang}]{zhao2024llama}
Jun Zhao, Zhihao Zhang, Qi~Zhang, Tao Gui, and Xuanjing Huang. 2024.
\newblock Llama beyond english: An empirical study on language capability transfer.
\newblock \emph{arXiv preprint arXiv:2401.01055}.

\end{thebibliography}

\clearpage
\newpage
\appendix

\setcounter{table}{0}
\setcounter{figure}{0}
\renewcommand{\thetable}{A\arabic{table}}
\renewcommand{\thefigure}{A\arabic{figure}}

\onecolumn
\section{Appendix}
\label{sec:appendix}

\subsection{Extended Results}
\label{sec:extended_results}

\begin{figure*}[ht]
\begin{tcolorbox}[width=\linewidth, sharp corners=all, colback=white!95!black]
\textbf{Prompt:} que partes tiene una noticia?

\textbf{Completion:} Una noticia típica suele tener las siguientes partes: ...

\textbf{Prompt:} y los epigrafes?

\textbf{Completion:} ... Los epígraf\begin{CJK}{UTF8}{min}\textcolor{red}{瓦解}\end{CJK} también se pueden utilizar para resaltar citas importantes, proporcionar transiciones entre temas o simplemente dividir el texto en secciones más manejables y digeribles.
\end{tcolorbox} 
\caption{Example of non-English word-level language confusion produced by an LLM.}
\label{fig:non_en_word_switch}
\end{figure*}

\begin{table*}[ht]
\resizebox{\textwidth}{!}{%
\centering
\begin{tabular}{@{}laccccccccccccccc@{}}
\toprule
              & \multicolumn{16}{c}{Monolingual}                                                                                                                                                                                                                                \\ \midrule
              & avg           & ar             & de             & en             & es            & fr             & hi             & id   & it             & ja             & ko             & pt            & ru             & tr             & vi             & zh            \\ \midrule
Llama 2 70B-I       & 49.8          & 3.2            & 60.3           & 99.7           & 96.7          & 89.8           & 2.3            & 65.6 & 73.0           & 7.3            & 0.7            & 92.5          & 91.2           & 35.5           & 17.7           & 11.7          \\
Llama 3 70B-I       & 47.3          & 27.7           & 31.0           & \textbf{100.0} & 98.8          & 89.1           & 25.4           & 21.5 & 88.9           & 11.2           & 2.3            & 97.8          & 77.0           & 19.6           & 10.4           & 9.4           \\
Mixtral 8x7B  & 74.2          & 49.5           & 90.9           & 99.8           & 90.0          & 95.3           & 72.2           & 60.2 & 72.0           & 70.6           & 62.1           & 86.2          & 65.7           & 90.0           & 59.4           & 49.2          \\
Mistral Large & 70.3          & 48.7           & 98.0           & 99.6           & 99.5          & \textbf{100.0} & 19.5           & 31.0 & 99.0           & 48.0           & 64.3           & 80.9          & 98.7           & 71.5           & 29.0           & 67.0          \\
Command R     & 99.5          & 100.0          & 99.3           & 99.9           & 98.7          & 99.9           & 100.0          & 96.8 & 99.7           & 100.0          & 100.0          & 99.7          & 100.0          & 99.0           & \textbf{99.9}  & 99.6          \\
Command R+    & \textbf{99.8} & 99.9           & \textbf{100.0} & \textbf{100.0} & 99.8          & 99.9           & \textbf{100.0} & 99.0 & \textbf{100.0} & 99.8           & \textbf{100.0} & 99.4          & \textbf{100.0} & \textbf{100.0} & 99.5           & 99.1          \\
GPT-3.5 Turbo & 99.5          & \textbf{100.0} & \textbf{100.0} & 99.8           & 99.8          & \textbf{100.0} & 99.5           & 98.3 & \textbf{100.0} & 98.7           & \textbf{100.0} & \textbf{99.5} & \textbf{100.0} & \textbf{100.0} & 99.0           & 98.3          \\
GPT-4 Turbo   & 99.7          & 99.2           & \textbf{100.0} & \textbf{100.0} & \textbf{99.8} & 99.7           & \textbf{100.0} & 98.0 & 99.7           & \textbf{100.0} & \textbf{100.0} & 99.4          & \textbf{100.0} & \textbf{100.0} & \textbf{100.0} & \textbf{99.7} \\ 
\bottomrule
\end{tabular}
}
\caption{\textbf{Line-level language ID accuracy of different LLMs on \textit{monolingual} generation.} Line accuracy is calculated by $\frac{\text{\# lines responded in correct language}}{\text{Total \# of lines generated}}$ }
\label{tab:monolingual-line-acc}
\end{table*}

\begin{table*}[ht]
\resizebox{\textwidth}{!}{%
\centering
\begin{tabular}{@{}lacccccccccccccc@{}}
\toprule
              & \multicolumn{15}{c}{Cross-lingual}                                                                                                                                                                                                            \\ \midrule
              & avg           & ar            & de            & es            & fr            & hi            & id            & it            & ja            & ko            & pt            & ru            & tr            & vi            & zh            \\ \midrule
Llama 2 70B-I       & 46.0          & 23.5          & 58.3          & 80.5          & 74.6          & 31.4          & 51.1          & 73.4          & 26.9          & 15.9          & 79.9          & 47.0          & 28.1          & 30.0          & 23.3          \\
Llama 3 70B-I       & 53.4          & 60.3          & 65.8          & 77.4          & 73.8          & 69.7          & 58.2          & 65.5          & 6.2           & 2.8           & 75.8          & 65.2          & 55.2          & 61.4          & 10.4          \\
Mixtral 8x7B  & 72.7          & 61.5          & 79.7          & 82.5          & 82.5          & 41.7          & 78.9          & 85.9          & 64.1          & 61.3          & 83.0          & 75.2          & 78.1          & 79.4          & 64.6          \\
Mistral Large & 65.7          & 45.4          & 79.8          & 74.9          & 76.7          & 63.3          & 68.6          & 73.5          & 58.2          & 51.9          & 72.4          & 68.7          & 69.5          & 61.4          & 56.1          \\
Command R     & 75.7          & 67.5          & 75.4          & 80.0          & 81.7          & 69.7          & 78.9          & 74.1          & 75.4          & 76.3          & 73.2          & 76.0          & 76.3          & 72.3          & 83.1          \\
Command R+    & \textbf{95.3} & \textbf{95.9} & \textbf{95.5} & \textbf{95.8} & \textbf{95.3} & \textbf{94.0} & \textbf{93.2} & \textbf{95.9} & \textbf{96.4} & \textbf{95.4} & \textbf{94.5} & \textbf{96.1} & \textbf{94.9} & \textbf{95.8} & \textbf{95.0} \\
GPT-3.5 Turbo & 91.5          & 91.5          & 91.5          & 94.9          & 90.9          & 92.2          & 88.0          & 92.7          & 89.7          & 91.3          & 93.9          & 92.6          & 90.5          & 92.5          & 89.4          \\
GPT-4 Turbo   & 92.4          & 91.0          & 94.0          & 94.7          & 93.1          & 92.6          & 92.0          & 93.2          & 90.5          & 91.5          & 94.4          & 92.0          & 92.4          & 91.5          & 90.3          \\ 
\bottomrule
\end{tabular}
} 
\caption{\textbf{Line-level language ID accuracy of different LLMs on \textit{cross-lingual} generation.} Line accuracy is calculated by $\frac{\text{\# lines responded in correct language}}{\text{Total \# of lines generated}}$ }
\label{tab:crosslingual-line-acc}
\end{table*}

\begin{table*}[ht]
\centering
\resizebox{\linewidth}{!}{%
\begin{tabular}{@{}laccccccacccccc@{}}
\toprule
\multicolumn{1}{l}{} & \multicolumn{7}{c}{\textbf{Monolingual}}                                                                                                 & \multicolumn{7}{c}{\textbf{Cross-lingual}}                                                                     \\ 
\multicolumn{1}{l}{} & Avg           & ar             & hi             & ja             & ko             & ru             & \multicolumn{1}{c|}{zh}             & Avg           & ar            & hi            & ja            & ko             & ru            & zh            \\ \midrule
Llama 2 70B-I        & 97.9          & \textbf{100.0} & \textbf{100.0} & \textbf{100.0} & \textbf{100.0} & 93.2           & \multicolumn{1}{c|}{94.4}           & 84.2          & 91.4          & 85.3          & 84.4          & 86.7           & 81.5          & 75.9          \\
Llama 3 70B-I        & 93.0          & 95.6           & 95.7           & 80.0           & \textbf{100.0} & 93.5           & \multicolumn{1}{c|}{93.3}           & 94.4          & 95.7          & 97.7          & 91.7          & \textbf{100.0} & 89.1          & 92.3          \\
Llama 3.1 70B-I & 99.5 & \textbf{100.0} & \textbf{100.0} & 99.0 & \textbf{100.0} & 98.0 & \multicolumn{1}{c|}{\textbf{100.0}} & 95.0 & 95.7 & 98.5 & 90.3 & 97.4 & 93.4 & 94.4 \\
Mixtral 8x7B         & 73.7          & 86.0           & 78.9           & 68.2           & 61.7           & 83.1           & \multicolumn{1}{c|}{64.7}           & 68.2          & 76.3          & 71.5          & 51.5          & 67.3           & 80.5          & 62.1          \\
Mistral Large        & 98.4          & \textbf{100.0} & 94.7           & 97.9           & \textbf{100.0} & 99.0           & \multicolumn{1}{c|}{98.9}           & 93.8          & 93.5          & 95.4          & 91.5          & 92.0           & 92.8          & 97.7          \\
Command R            & 96.3          & 99.3           & 99.0           & 93.9           & 97.0           & 96.0           & \multicolumn{1}{c|}{92.3}           & 94.0          & 94.3          & 98.6          & 88.5          & 97.2           & 94.0          & 91.1          \\
Command R+           & 99.4          & 99.7           & \textbf{100.0} & 99.0           & \textbf{100.0} & 98.0           & \multicolumn{1}{c|}{\textbf{100.0}} & 95.1          & 97.9          & 96.0          & 95.5          & 96.1           & 89.7          & 95.6          \\
Command R Refresh  & 99.4 & 99.7  & \textbf{100.0} & 99.0  & \textbf{100.0} & 99.0  & \multicolumn{1}{c|}{99.0} & 97.2 & 97.0 & 98.8 & 96.7 & 96.6  & 95.5 & 98.5 \\
Command R+ Refresh & \textbf{99.8} & 99.0  & \textbf{100.0} & \textbf{100.0} & \textbf{100.0} & \textbf{100.0} & \multicolumn{1}{c|}{\textbf{100.0}} & 96.5 & 97.3 & 97.8 & 96.6 & 96.9  & 94.5 & 95.7 \\
GPT-3.5 Turbo        & \textbf{99.8} & \textbf{100.0} & \textbf{100.0} & 99.0           & \textbf{100.0} & \textbf{100.0} & \multicolumn{1}{c|}{\textbf{100.0}} & \textbf{98.7} & 99.0 & \textbf{99.1} & \textbf{98.6} & 98.5           & \textbf{98.3} & \textbf{99.0} \\
GPT-4 Turbo          & 99.7          & \textbf{100.0} & \textbf{100.0} & \textbf{100.0} & 99.0           & \textbf{100.0} & \multicolumn{1}{c|}{99.0}           & 96.6          & 97.4          & 97.3          & 95.6          & 96.8           & 95.7          & 97.1          \\
GPT-4o             & 99.7 & \textbf{100.0} & \textbf{100.0} & \textbf{100.0} & 99.0  & 99.0  & \multicolumn{1}{c|}{\textbf{100.0}} & 98.1 & \textbf{99.2} & 97.0 & 97.5 & 99.1  & 98.0 & 97.6 \\
\bottomrule
\end{tabular}
}
\caption{\textbf{Word-level pass rate (WPR) on \textit{monolingual} and \textit{cross-lingual} generation in non-Latin script languages} (\% of responses containing no English words). Llama models are instruction-tuned variants.}
\label{tab:lang-word-level}
\end{table*}

\begin{table}[]
\resizebox{\textwidth}{!}{%
\begin{tabular}{lccccccccccacccccccc}
\toprule
& \multicolumn{10}{c}{\textbf{Monolingual}} 
& \multicolumn{9}{c}{\textbf{Cross-lingual}}                                                                     \\
& Avg   & de    & en    & es    & fr    & id    & it    & pt    & tr    & vi  & Avg  & de    & es    & fr    & id    & it    & pt    & tr    & vi    \\ \midrule
Llama 3 70B-I      & 99.0  & \textbf{100.0} & 99.7  & \textbf{100.0} & \textbf{100.0} & \textbf{100.0} & \textbf{100.0} & \textbf{100.0} & \textbf{100.0} & \multicolumn{1}{c|}{90.9} & 99.7 & \textbf{100.0} & 99.0  & \textbf{100.0} & \textbf{100.0} & \textbf{100.0} & \textbf{100.0} & \textbf{100.0} & 98.7 \\
Llama 3.1 70B-I    & 99.6  & \textbf{100.0} & 99.7  & \textbf{100.0} & \textbf{100.0} & 98.9  & \textbf{100.0} & \textbf{100.0} & 99.0  & \multicolumn{1}{c|}{99.0} & 99.1 & \textbf{100.0} & 99.2  & 99.6  & 99.1  & \textbf{100.0} & 99.6  & 98.0  & 97.5 \\
Command R          & 99.2  & \textbf{100.0} & 99.7  & \textbf{100.0} & \textbf{100.0} & \textbf{100.0} & \textbf{100.0} & \textbf{100.0} & 96.9  & \multicolumn{1}{c|}{96.0} & 97.9 & 98.9  & \textbf{100.0} & \textbf{100.0} & 96.4  & \textbf{100.0} & 98.5  & 97.4  & 92.2  \\
Command R+         & 99.6  & \textbf{100.0} & \textbf{100.0} & \textbf{100.0} & \textbf{100.0} & 99.0  & \textbf{100.0} & \textbf{100.0} & \textbf{100.0} & \multicolumn{1}{c|}{97.0} & 98.6 & \textbf{100.0} & \textbf{100.0} & \textbf{100.0} & 98.6  & 99.6  & \textbf{100.0} & 97.9  & 93.1   \\
Command R Refresh  & 99.9  & \textbf{100.0} & \textbf{100.0} & 99.7  & \textbf{100.0} & \textbf{100.0} & \textbf{100.0} & 99.5  & \textbf{100.0} & \multicolumn{1}{c|}{100.0} & \textbf{99.9} & \textbf{100.0} & \textbf{100.0} & \textbf{100.0} & 99.5  & \textbf{100.0} & \textbf{100.0} & \textbf{100.0} & \textbf{100.0} \\
Command R+ Refresh & 99.9  & \textbf{100.0} & 99.9  & \textbf{100.0} & \textbf{100.0} & \textbf{100.0} & \textbf{100.0} & \textbf{100.0} & \textbf{100.0} & \multicolumn{1}{c|}{99.0} & 99.7 & \textbf{100.0} & \textbf{100.0} & \textbf{100.0} & 99.1  & \textbf{100.0} & \textbf{100.0} & 99.0  & 99.6  \\
GPT-3.5 Turbo      & \textbf{100.0} & \textbf{100.0} & 99.9  & \textbf{100.0} & \textbf{100.0} & \textbf{100.0} & \textbf{100.0} & \textbf{100.0} & \textbf{100.0} & \multicolumn{1}{c|}{\textbf{100.0}} & \textbf{99.9} & \textbf{100.0} & \textbf{100.0} & \textbf{100.0} & \textbf{100.0} & \textbf{100.0} & \textbf{100.0} & 99.2  & \textbf{100.0} \\
GPT-4o             & 99.8  & \textbf{100.0} & 99.7  & \textbf{100.0} & \textbf{100.0} & 98.9  & \textbf{100.0} & 99.5  & \textbf{100.0} & \multicolumn{1}{c|}{\textbf{100.0}} & 99.7 & \textbf{100.0} & \textbf{100.0} & \textbf{100.0} & 99.1  & \textbf{100.0} & \textbf{100.0} & 99.6  & 99.1  \\
\bottomrule
\end{tabular}%
}
\caption{\textbf{Word-level pass rate (WPR) on \textit{monolingual} and \textit{cross-lingual} generation in Latin script languages} (\% of responses containing no characters that are outside the corresponding script's Unicode range). Llama models are instruction-tuned variants.}
\label{tab:lang-word-level-latin-script}
\end{table}

\begin{table*}[ht]
\centering
\resizebox{0.95\textwidth}{!}{%
\begin{tabular}{@{}laccccccacccccc@{}}
\toprule
\multicolumn{1}{l}{} & \multicolumn{7}{c}{\textbf{Monolingual}}                                                                                                         & \multicolumn{7}{c}{\textbf{Cross-lingual}}                                                                             \\ 
\multicolumn{1}{l}{} & Avg                           & ar                   & hi                   & ja                   & ko                   & ru                   & \multicolumn{1}{c|}{zh}            & Avg           & ar            & hi            & ja            & ko            & ru            & zh            \\ \midrule
Llama 2 70B-I        & 20.9                          & 0.7                  & 2.0                  & 13.1                 & 0.0                  & 91.0                 & \multicolumn{1}{c|}{18.9}          & 27.2          & 21.8          & 33.9          & 27.2          & 9.0           & 52.1          & 19.3          \\
Llama 3 70B-I        & 31.6                          & 35.3                 & 37.1                 & 17.8                 & 0.0                  & 84.5                 & \multicolumn{1}{c|}{14.7}          & 29.3          & 46.9          & 62.9          & 2.7           & 1.5           & 53.7          & 8.3           \\
Llama 3.1 70B-I & 99.3 & 99.4 & \textbf{100.0} & 97.9 & \textbf{100.0} & 99.0 & \multicolumn{1}{c|}{\textbf{99.5}} & 81.0 & 85.5 & 97.0 & 71.7 & 67.4 & 84.1 & 80.3 \\
Mixtral 8x7B         & 65.3                          & 61.9                 & 74.7                 & 67.4                 & 61.4                 & 72.9                 & \multicolumn{1}{c|}{53.4}          & 61.6          & 66.6          & 50.6          & 54.5          & 61.7          & 76.2          & 59.7          \\
Mistral Large        & 69.4                          & 64.9                 & 31.7                 & 64.4                 & 78.0                 & 98.5                 & \multicolumn{1}{c|}{79.2}          & 63.3          & 52.1          & 72.5          & 59.9          & 56.9          & 75.3          & 63.0          \\
Command R            & 98.0                          & 99.7                 & 99.5                 & 96.9                 & 98.5                 & 98.0                 & \multicolumn{1}{c|}{95.3}          & 78.6          & 74.5          & 78.7          & 75.1          & 80.9          & 79.9          & 82.3          \\
Command R+           & 99.4                          & 99.7                 & \textbf{100.0}       & 99.0                 & \textbf{100.0}       & 99.0                 & \multicolumn{1}{c|}{98.7}          & 93.4          & 95.6 & 93.0          & 94.6 & 93.5          & 91.3          & 92.5 \\
Command R Refresh  & 99.4 & 99.6  & \textbf{100.0} & 99.0  & \textbf{100.0} & 99.5  & \multicolumn{1}{c|}{98.5} & 94.9 & 94.4 & 96.9 & 95.8 & 95.2 & 93.8 & \textbf{93.5} \\
Command R+ Refresh & 99.6 & 99.0  & \textbf{100.0} & \textbf{100.0} & \textbf{100.0} & 99.5  & \multicolumn{1}{c|}{99.0} & \textbf{95.8} & 96.4 & \textbf{98.0} & 95.8 & \textbf{96.4} & \textbf{95.4} & 92.9 \\
GPT-3.5 Turbo        & 99.4                          & \textbf{100.0}       & 99.5                 & 98.5                 & \textbf{100.0}       & \textbf{100.0}       & \multicolumn{1}{c|}{98.5}          & 94.1 & 94.7          & 95.4 & 93.2          & 94.2 & 94.9 & 92.3          \\
GPT-4 Turbo          & \textbf{99.7}                 & 99.5                 & \textbf{100.0}       & \textbf{100.0}       & 99.5                 & \textbf{100.0}       & \multicolumn{1}{c|}{99.0} & 92.8          & 93.0          & 94.1          & 91.5          & 93.1          & 92.8          & 92.3          \\ 
GPT-4o             & 99.6 & 99.8  & \textbf{100.0} & \textbf{100.0} & 99.2  & 99.2  & \multicolumn{1}{c|}{99.5} & 95.1 & \textbf{97.0} & 94.9 & \textbf{96.7} & 96.2 & 94.9 & 91.1 \\
\bottomrule
\end{tabular}%
}
\caption{\textbf{Aggregate language confusion pass rate (LCPR) of LLMs on \textit{monolingual} and \textit{cross-lingual} tasks}.} 
\label{tab:lang-lcpr}
\end{table*}

\clearpage
\newpage
\onecolumn
\begin{table*}[h]
\centering
\begin{tabular}{@{}cp{3cm}p{7cm}l@{}}
\toprule
models        & \multicolumn{1}{c}{prompt}                                            & \multicolumn{1}{c}{generation}                                                                                                                                                                                                                                                                                                                                                                                                                                                                                                                                                                                                                            & \multicolumn{1}{c}{confusion type} \\ \midrule
Llama 3 70B-I & \begin{CJK*}{UTF8}{gbsn}ローイングエクササイズの効果について論じる記事を1,000文字以内で書いてください。\end{CJK*}                            & \textcolor{red}{**The Effects of Rowing Exercise: A Comprehensive Review**\textbackslash{}n\textbackslash{}nRowing exercise has gained popularity in recent years due to its numerous benefits for physical and mental health. As a low-impact, full-body workout, rowing has been shown to improve cardiovascular fitness, increase muscle strength and endurance, and enhance overall well-being. In this article, we will discuss the effects of rowing exercise and explore its advantages for individuals of all fitness levels.\textbackslash{}n\textbackslash{}n**Cardiovascular Benefits**\textbackslash{}n\textbackslash{}nRowing is an excellent cardiovascular} & Full response                      \\
Mixtral 8x7B  & \begin{CJK*}{UTF8}{gbsn}问题：油在生活中有哪些用途？\end{CJK*}                                                        & \begin{CJK*}{UTF8}{gbsn}'油在我们的日常生活中有许多用途，主要包括：\textbackslash{}n\textbackslash{}n1. 烹饪：油是烹饪中不可或缺的一种原料，它可以用来油炸、煎炸、烧烤等 \end{CJK*} \textcolor{red}{verschiedenen Kochtechniken. Es kann auch als Salatöl oder Dressing verwendet werden.}\textbackslash{}n'                                                                                                                                                                                                                                                                                                                                                                                                                                                       & Line level                         \\
Command R     & \begin{CJK}{UTF8}{mj}교실에서 시작하는 디지털 시민성 교육에 관한 교육 자료를 초등학교 수준으로 쉬운 말투를 사용하여 200단어 이내로 작성해줘.\end{CJK} & \begin{CJK}{UTF8}{mj}디지털 시민이란 인터넷과 디지털 기술로 연결된 세상에서 다른 사람들과 상호작용하고, 소통하는 방법을 아는 사람을 말해. 디지털 세상에서 우리는 좋은 친구들이나 가족들처럼 행동해야 해. \textbackslash{}n\textbackslash{}n교실에서 디지털 시민이 되는 법을 배워보아요! 우리는 먼저 인터넷에서 주의해야 할 점을 알아볼 거야. 디지털 세상에서 우리\end{CJK}  \textcolor{red}{would} \begin{CJK}{UTF8}{mj}안전한 웹사이트만 방문하고, 우리 개인정보를 소중히 지켜야 해.\end{CJK}                                                                                                                                                                                                                                                                                                                                                                                                  & Word level                         \\ \bottomrule
\end{tabular}
\caption{\textbf{Examples of different types of language confusion across models.}}
\label{fig:lang-confuse-examples}
\end{table*}

\clearpage
\newpage

\twocolumn

\subsection{Annotator statistics} \label{sec:app-annotator-stats}

Prompts were written by native-level speakers of the respective languages, each of whom is also fluent in English.
Annotators were paid by the hour, with compensation above the federal minimum wage of the country of employment.

\subsection{Cross-lingual prompt generation} \label{sec:app-cross-lingual-prompt-generation}

In order to generate prompts for the cross-lingual setting, we semi-automatically amend English prompts with an English instruction to generate text in the target language. For each English prompt and target language, we randomly choose an English instruction (e.g., ``Respond in French'', ``Reply in Turkish''). To control for the position of the instruction in the prompt, for each prompt separate examples are created where an instruction is inserted at the beginning and at the end respectively. For each prompt, we additionally generate another example where an instruction is manually integrated into the prompt, e.g., ``Generate an essay \textbf{in Korean}''. The same process is applied to Okapi, ShareGPT, and our complex prompts. Table \ref{tab:example-prompts} shows representative examples where instructions are integrated in the middle, inserted at the beginning and at the end of the original prompts.

\clearpage

\subsection{Dataset Language Confusion Metrics}
\label{sec:dataset_lang_confuse}

\begin{table}[htb]
\resizebox{\linewidth}{!}{%
\begin{tabular}{@{}lacccc@{}}
\toprule
 & Avg & Okapi & Aya & Dolly & Native (Ours) \\ \midrule
Llama 2 70B-I & 54.0 & 57.2 & 49.6 & 55.2 & 49.0 \\
Llama 3 70B-I & 56.2 & 56.0 & 50.8 & 61.8 & 47.0 \\
Llama 3.1 70B-I & \textbf{99.7} & \textbf{100.0} & \textbf{100.0} & 99.3 & 99.5 \\
Mixtral 8x7B & 74.9 & 70.7 & 78.3 & 75.8 & 78.2 \\
Mistral Large & 75.6 & 68.4 & 82.0 & 76.4 & 77.8 \\
Command R & 98.7 & 97.8 & 99.4 & 99.0 & 98.8 \\
Command R+ & 99.3 & 99.0 & 99.0 & 99.8 & 99.2 \\
Command R Refresh  & 99.2 & 98.4  & 99.0 & \textbf{100.0} & 99.5 \\
Command R+ Refresh & 99.2 & 99.2  & 98.6 & 99.4  & \textbf{99.8} \\
GPT-3.5 Turbo & 99.1 & 99.0 & 98.6 & 99.8 & 99.5 \\
GPT-4 Turbo & 99.4 & 98.9 & 99.4 & 99.8 & 99.2 \\
GPT-4o             & 99.1 & 98.8  & 98.8 & 99.2  & 99.5 \\
\bottomrule
\end{tabular}%
}
\caption{\textbf{Line-level pass rate (LPR) by dataset on monolingual generation.}}
\label{tab:dataset-impact-lpr-monolingual}
\end{table}

\begin{table*}[]
\centering
\resizebox{\textwidth}{!}{%
\begin{tabular}{@{}laccccaccc@{}}
\toprule
\multicolumn{1}{l}{} & \multicolumn{5}{c}{\textbf{Monolingual}}                                                                    & \multicolumn{4}{c}{\textbf{Cross-lingual}}                              \\ 
\multicolumn{1}{l}{} & Avg           & Okapi         & Aya           & Dolly         & \multicolumn{1}{c|}{Native (Ours)} & Avg           & Okapi         & ShareGPT      & Complex (Ours) \\ \midrule
Llama 2 70B-I        & 97.4                 & \textbf{100.0} & 94.4                 & 97.7                 & \multicolumn{1}{c|}{\textbf{100.0}} & 84.2                 & 90.0           & 78.9                 & 83.7                 \\
Llama 3 70B-I        & 95.1                 & \textbf{100.0} & 93.3                 & 91.9                 & \multicolumn{1}{c|}{90.0}           & 94.4                 & 98.9           & 87.9                 & 96.4        \\
Llama 3.1 70B-I & 99.7 & \textbf{100.0} & \textbf{100.0} & 99.3 & \multicolumn{1}{c|}{99.5} & 95.0 & 97.7 & 93.3 & 94.0 \\
Mixtral 8x7B         & 78.0                 & 77.9           & 74.8                 & 81.3                 & \multicolumn{1}{c|}{64.9}           & 68.2                 & 75.8           & 68.0                 & 60.8                 \\
Mistral Large        & 98.9                 & \textbf{100.0} & 98.9                 & 97.9                 & \multicolumn{1}{c|}{99.0}           & 93.8                 & 98.7           & 95.4                 & 87.3                 \\
Command R            & 96.5                 & 97.4           & 93.8                 & 98.3                 & \multicolumn{1}{c|}{95.5}           & 94.0                 & 95.3           & 93.7                 & 92.9                 \\
Command R+           & 99.7                 & \textbf{100.0} & \textbf{100.0}       & 99.0                 & \multicolumn{1}{c|}{99.5}           & 95.1                 & 97.4           & 97.6                 & 90.3                 \\
Command R Refresh           & 99.5                 & \textbf{100.0} & 99.0       & 99.3                 & \multicolumn{1}{c|}{99.5}           & 98.8                 & 99.6           & 99.6                 & 97.1                 \\
Command R+ Refresh           & 99.6                 & \textbf{100.0} & 99.5       & 99.0                 & \multicolumn{1}{c|}{\textbf{100.0}}           & 98.3                & 99.3           & 99.1                 & 96.6                 \\
GPT-3.5 Turbo        & \textbf{100.0}       & \textbf{100.0} & \textbf{100.0}       & \textbf{100.0}       & \multicolumn{1}{c|}{99.5}           & 98.7        & \textbf{100.0} & \textbf{99.8}        & 96.4        \\
GPT-4 Turbo          & 99.7                 & \textbf{100.0} & 99.0                 & \textbf{100.0}       & \multicolumn{1}{c|}{99.5}           & 96.6                 & 99.8           & 99.5                 & 90.6                 \\ 
GPT-4o          & 99.8                 & \textbf{100.0} & \textbf{100.0}                 & 99.5       & \multicolumn{1}{c|}{99.7}           & \textbf{99.0}                 & \textbf{100.0}           & 99.6                 & \textbf{97.4}                 \\ \bottomrule
\end{tabular}
}
\caption{\textbf{WPR by dataset for \textit{monolingual} and \textit{cross-lingual} generation.}}
\label{tab:dataset-impact-wpr}
\end{table*}

\begin{table*}[]
\centering
\resizebox{\textwidth}{!}{%
\begin{tabular}{@{}laccccaccc@{}}
\toprule
\multicolumn{1}{l}{} & \multicolumn{5}{c}{\textbf{Monolingual}}                                                                    & \multicolumn{4}{c}{\textbf{Cross-lingual}}                              \\ 
\multicolumn{1}{l}{} & Avg           & Okapi         & Aya           & Dolly         & \multicolumn{1}{c|}{Native (Ours)} & Avg           & Okapi         & ShareGPT      & Complex (Ours) \\ \midrule
Llama 2 70B-I        & 69.4          & 72.8          & 65.0          & 70.5          & \multicolumn{1}{c|}{65.8}          & 52.0          & 58.6          & 58.1          & 39.3           \\
Llama 3 70B-I        & 70.5          & 71.8          & 65.8          & 73.9          & \multicolumn{1}{c|}{61.8}          & 44.7          & 51.7          & 54.5          & 27.9           \\
Llama 3.1 70B-I & 99.4 & 99.5 & 99.3 & 99.4 & \multicolumn{1}{c|}{99.3} & 87.3 & 91.1 & 92.1 & 78.7 \\
Mixtral 8x7B         & 76.4          & 74.1          & 76.5          & 78.5          & \multicolumn{1}{c|}{71.0}          & 68.3          & 76.6          & 73.4          & 54.8           \\
Mistral Large        & 85.6          & 81.2          & 89.7          & 85.8          & \multicolumn{1}{c|}{87.1}          & 71.7          & 80.5          & 70.8          & 64.0           \\
Command R            & 97.6          & 97.6          & 96.5          & 98.7          & \multicolumn{1}{c|}{97.1}          & 76.9          & 84.3          & 91.6          & 54.9           \\
Command R+           & 99.5          & 99.5 & \textbf{99.5} & 99.4          & \multicolumn{1}{c|}{99.4}          & 93.0          & 96.7          & 98.2          & 84.2           \\
Command R Refresh           & 99.4          & 99.2          & 99.0          & 99.7         & \multicolumn{1}{c|}{99.5}          & 95.7          & \textbf{98.9}          & \textbf{99.5}          & 88.7           \\
Command R+ Refresh          & 99.4          & \textbf{99.6} & 99.0 & 99.2          & \multicolumn{1}{c|}{\textbf{99.9}}          & \textbf{96.8}          & 98.6          & 98.9          & \textbf{93.0}           \\
GPT-3.5 Turbo        & \textbf{99.6} & 99.5 & 99.3          & \textbf{99.9} & \multicolumn{1}{c|}{99.5} & 93.8 & 98.8 & 98.3 & 84.4  \\
GPT-4 Turbo          & 99.5          & 99.4          & 99.2          & \textbf{99.9} & \multicolumn{1}{c|}{99.4}          & 93.3          & 98.2          & 97.9          & 83.7           \\ 
GPT-4o          & 99.4          & 99.4          & 99.4          & 99.3 & \multicolumn{1}{c|}{99.6}          & 95.5          & 98.7          & 98.6          & 89.1           \\ \bottomrule
\end{tabular}
}
\caption{\textbf{LCPR by dataset for \textit{monolingual} and \textit{cross-lingual} generation.}}
\label{tab:dataset-impact-lcpr}
\end{table*}

\clearpage
\newpage
\twocolumn
\subsection{Extended Impacts}
\label{sec:extended_impacts}

\begin{table}[ht]
\centering
\resizebox{\linewidth}{!}{%
\begin{tabular}{lccc}
\toprule
 & Short & Medium & Long \\
 & [21, 65] & [68, 224] & [227, 2971] \\
 \midrule
Command R & 42.7 & 28.1 & 38.0 \\
\quad + one-shot & 59.5 & 49.1 & 67.7 \\
Command R+ & 82.1 & 74.8 & 79.8 \\
GPT-4 & 89.3 & 72.9 & 71.8 \\
\bottomrule
\end{tabular}
}
\caption{\textbf{Line-level pass rate (LPR) on our ``Complex prompts'' dataset for cross-lingual generation, depending on length of the prompts.} We sort the prompts by length and split them into 3 length buckets of the same size (each containing one third of the prompts). [a, b]: min and max length in words of each bucket's prompts.}
\label{tab:impact-prompt-length}
\end{table}

\begin{table}[ht]
\centering
\begin{tabular}{lccc}
\toprule
 & Start & Integrated & End \\
 \midrule
Command R & 86.7 & 69.0 & 85.1 \\
\quad + one-shot & 88.7 & 80.6 & 90.6 \\
Command R+ & 94.4 & 90.3 & 95.2 \\
GPT-4 & 93.0 & 91.7 & 95.0 \\
\bottomrule
\end{tabular}
\caption{\textbf{Line-level pass rate (LPR) on crosslingual generation depending on the position of the language control instruction.} ``Integrated'' corresponds to instructions of the form ``Write an essay of 100 words in Korean about artificial intelligence.'' ``Start'' and ``End'' are isolated instructions of the form ``Reply in French.'' placed either at the start or at the end of the prompt.}
\label{tab:impact-instruction-position}
\end{table}

\newpage
\subsection{Quantization}
\label{sec:app-quant}
It is common to train at half-precision floating-point (FP16), where weights and activations of a network use 16 bits (2 bytes) to represent a floating-point value. It is common to quantize weights to INT8 (8-bit integers), often referred to as \textit{W8}.  More extreme quantization of weights to INT4 (4-bit) is called \textit{W4}. Quantizing both weights and activations to INT8 is commonly called \textit{W8A8}.
In Section \ref{sec:quant}, we compare \textit{FP16} with \textit{W8}, \textit{W8A8}, and \textit{W4}\footnote{W4 with group-wise scaling using GPTQ \citep{frantar-gptq}.} variants of Command R+ on monolingual generation.

\begin{table}[ht]
\centering\footnotesize
\resizebox{\linewidth}{!}{
	\addtolength{\tabcolsep}{-3pt}
    \begin{tabular}{@{}larrrrrrrr@{}}
    \toprule
     & \multicolumn{8}{c}{Line-level Pass Rate (LPR)} \\
     & \multicolumn{1}{c}{avg} & \multicolumn{1}{c}{ar} & \multicolumn{1}{c}{hi} & \multicolumn{1}{c}{ja} & \multicolumn{1}{c}{ko} & \multicolumn{1}{c}{ru} & \multicolumn{1}{c}{zh} & \multicolumn{1}{c}{id} \\
     \cmidrule{2-9}
    \textbf{FP16} & 99.3 & 99.0 & 100.0 & 99.0 & 100.0 & 100.0 & 97.5 & 97.0 \\
    \textbf{W8} & 99.3 & 99.7 & 100.0 & 98.0 & 100.0 & 100.0 & 99.0 & 97.0 \\
    \textbf{W8A8} & 99.5 & 100.0 & 100.0 & 100.0 & 100.0 & 100.0 & 98.5 & 98.0 \\
    \textbf{W4-g} & 98.8 & 99.3 & { 99.0} & 98.0 & 100.0 & 100.0 & 98.5 & { 91.0} \\
     \toprule
     & \multicolumn{8}{c}{Word-level Pass Rate (WPR)} \\
     \cmidrule{2-9}
    \textbf{FP16} & 98.2 & 99.3 & 100.0 & 100.0 & 99.0 & 95.0 & 96.0 & \multicolumn{1}{c}{-} \\
    \textbf{W8} & 98.3 & 99.3 & 99.0 & 99.0 & 100.0 & 96.0 & 96.5 & \multicolumn{1}{c}{-} \\
    \textbf{W8A8} & 98.7 & 99.3 & 99.0 & 100.0 & 100.0 & 97.0 & 97.0 & \multicolumn{1}{c}{-} \\
    \textbf{W4-g} & 98.1 & 99.3 & 100.0 & { 98.0} & 99.0 & { 93.9} & 98.5 & \multicolumn{1}{c}{-} \\ \bottomrule
    \end{tabular}
}
\caption{\textbf{Effect of quantization on LPR and WPR on monolingual generation.}}
\label{tab:quant-short}
\end{table}

\clearpage
\subsection{Extended Beam Search and Nucleus Sampling Results}
\label{sec:app-nuc-samp-beam-search-extended}

\begin{table}[ht]
\footnotesize
\centering
\begin{tabular}{@{}l|a|rrrrrr@{}}
\toprule
 & \multicolumn{1}{c}{avg} & \multicolumn{1}{c}{ar} & \multicolumn{1}{c}{hi} & \multicolumn{1}{c}{ja} & \multicolumn{1}{c}{ko} & \multicolumn{1}{c}{ru} & \multicolumn{1}{c}{zh} \\ \midrule
T=0.0  & 93.3          & \textbf{98.9} & 94.6          & 86.4          & 96.7          & 93.0          & 90.0          \\
T=0.3  & 94.0          & 94.3          & \textbf{98.6} & 88.5          & \textbf{97.2} & 94.0          & 91.1          \\
T=0.5  & 94.2          & \textbf{98.9} & 96.1          & 87.1          & 95.2          & \textbf{95.5} & \textbf{92.3} \\
T=0.7  & \textbf{94.6} & 97.9          & 99.1          & \textbf{90.8} & 95.7          & 94.6          & 89.6          \\
T=1.0  & 80.1          & 93.5          & 88.6          & 70.5          & 85.5          & 71.0          & 71.2          \\ \midrule
p=0.1  & 93.6          & 97.4          & 96.6          & \textbf{89.0} & 97.5          & 93.3          & 87.6          \\
p=0.3  & \textbf{94.0} & \textbf{98.0} & 97.6          & 85.7          & 98.4          & 92.0          & \textbf{92.3} \\
p=0.5  & \textbf{94.0} & 95.0          & 97.3          & \textbf{89.0} & \textbf{98.8} & 93.8          & 90.1          \\
p=0.75 & \textbf{94.0} & 94.3          & \textbf{98.6} & 88.5          & 97.2          & \textbf{94.0} & 91.1          \\ \bottomrule
\end{tabular}
\caption{\textbf{Effect of varying temperature ($T$) or nucleus size ($p$) on cross-lingual word-level language confusion (WPR) of \textit{Command R}}. Default values are $p=0.75$ and $T=0.3$. Best score is in \textbf{bold}.}
\label{tab:app-vary-pt-cross}
\end{table}

\begin{table}[htb]
\centering
\resizebox{\columnwidth}{!}{%
\begin{tabular}{@{}rarrrrrr@{}}
\toprule
\multicolumn{1}{l}{} &
  \multicolumn{7}{c}{\textbf{Monolingual}} \\
 &
  \multicolumn{1}{a}{\textbf{avg}} &
  \multicolumn{1}{c}{\textbf{ar}} &
  \multicolumn{1}{c}{\textbf{hi}} &
  \multicolumn{1}{c}{\textbf{ja}} &
  \multicolumn{1}{c}{\textbf{ko}} &
  \multicolumn{1}{c}{\textbf{ru}} & 
  \multicolumn{1}{c}{\textbf{zh}} \\  \midrule
1 &
  97.8 &
  98.6 &
  \textbf{100.0} &
  97.0 &
  97.0 &
  98.0 &
  96.4 \\
2 &
  98.6 &
  98.2 &
  \textbf{100.0} &
  98.0 &
  \textbf{100.0} &
  98.0 &
  97.4 \\
3 &
  98.6 &
  98.9 &
  \textbf{100.0} &
  \textbf{100.0} &
  98.0 &
  98.0 &
  96.9 \\
5 &
  \textbf{99.0} &
  \textbf{99.3} &
  \textbf{100.0} &
  99.0 &
  98.0 &
  \textbf{100.0} &
  97.4 \\
10 &
  \textbf{99.0} &
  98.5 &
  \textbf{100.0} &
  99.0 &
  98.0 &
  \textbf{100.0} &
  \textbf{98.4} \\ \bottomrule \toprule
\multicolumn{1}{l}{} &
  \multicolumn{7}{c}{\textbf{Cross-lingual}} \\
 &
  \multicolumn{1}{c}{\textbf{avg}} &
  \multicolumn{1}{c}{\textbf{ar}} &
  \multicolumn{1}{c}{\textbf{hi}} &
  \multicolumn{1}{c}{\textbf{ja}} &
  \multicolumn{1}{c}{\textbf{ko}} &
  \multicolumn{1}{c}{\textbf{ru}} &
  \multicolumn{1}{c}{\textbf{zh}} \\  \midrule
1 &
  94.9 &
  98.9 &
  97.2 &
  89.4 &
  97.5 &
  95.8 &
  90.4 \\
2 &
  95.4 &
  99.2 &
  97.1 &
  91.1 &
  95.7 &
  \textbf{97.2} &
  92.4 \\
3 &
  \textbf{97.1} &
  \textbf{100.0} &
  97.9 &
  95.7 &
  98.0 &
  96.4 &
  94.5 \\
5 &
  96.7 &
  99.7 &
  \textbf{98.3} &
  95.1 &
  96.4 &
  96.2 &
  \textbf{94.7} \\
10 &
  96.7 &
  98.3 &
  \textbf{98.3} &
  \textbf{96.7} &
  \textbf{98.4} &
  94.5 &
  93.7 \\ \bottomrule
\end{tabular}%
}
\caption{\textbf{\textit{Monolingual} and \textit{Cross-lingual} word-level pass rate (WPR) of \textit{Command R} using beam search} with beam sizes 1--10. \textit{($\neg$) IE = (non-) Indo-European language. ($\neg$) Latin = (non-) Latin script.} }
\label{tab:beam-search-full-wpr}
\end{table}

\begin{table*}[htb]
\centering
\resizebox{\textwidth}{!}{%
\begin{tabular}{@{}larrcrrrrrrrrrrrr|cccc@{}}
\toprule
\multicolumn{1}{l}{} & \multicolumn{20}{c}{\textbf{Monolingual}} \\
 & \multicolumn{1}{c}{\textbf{avg}} & \multicolumn{1}{c}{\textbf{ar}} & \multicolumn{1}{c}{\textbf{de}} & \textbf{en} & \multicolumn{1}{c}{\textbf{es}} & \multicolumn{1}{c}{\textbf{fr}} & \multicolumn{1}{c}{\textbf{hi}} & \multicolumn{1}{c}{\textbf{id}} & \multicolumn{1}{c}{\textbf{it}} & \multicolumn{1}{c}{\textbf{ja}} & \multicolumn{1}{c}{\textbf{ko}} & \multicolumn{1}{c}{\textbf{pt}} & \multicolumn{1}{c}{\textbf{ru}} & \multicolumn{1}{c}{\textbf{tr}} & \multicolumn{1}{c}{\textbf{vi}} & \multicolumn{1}{c}{\textbf{zh}} & \textbf{$\neg$ IE} & \textbf{IE} & \textbf{$\neg$ Latin} & \textbf{Latin} \\ \midrule
T=0.0 & 98.8 & \textbf{100.0} & \textbf{98.0} & \multicolumn{1}{c}{99.0} & \textbf{96.7} & 99.0 & \textbf{100.0} & 96.0 & 99.0 & 99.0 & \textbf{100.0} & 98.0 & \textbf{100.0} & 99.0 & \textbf{100.0} & \textbf{99.0} & 99.1 & 98.5 & \textbf{99.7} & 98.3 \\
T=0.3 & 98.6 & \textbf{100.0} & \textbf{98.0} & \multicolumn{1}{c}{\textbf{99.5}} & 95.7 & 99.3 & \textbf{100.0} & 92.0 & 99.0 & \textbf{100.0} & \textbf{100.0} & 98.5 & \textbf{100.0} & \textbf{99.0} & 99.0 & 98.5 & 98.6 & \textbf{98.6} & \textbf{99.7} & 97.8 \\
T=0.5 & 98.7 & 99.0 & \textbf{98.0} & \multicolumn{1}{c}{99.0} & 96.3 & \textbf{99.7} & \textbf{100.0} & 94.0 & 99.0 & \textbf{100.0} & \textbf{100.0} & 98.5 & \textbf{100.0} & 99.0 & 99.0 & 98.5 & 98.7 & \textbf{98.6} & 99.6 & 98.1 \\
T=0.7 & \textbf{98.9} & \textbf{100.0} & \textbf{98.0} & \multicolumn{1}{c}{99.0} & 95.7 & 99.0 & 99.0 & \textbf{99.0} & \textbf{100.0} & 99.0 & \textbf{100.0} & \textbf{99.0} & \textbf{100.0} & 99.0 & 99.0 & 98.5 & \textbf{99.3} & 98.5 & 99.4 & \textbf{98.6} \\
T=1.0 & 96.8 & \textbf{100.0} & 96.0 & \multicolumn{1}{c}{\textbf{99.5}} & 95.0 & \textbf{99.7} & 98.0 & 91.9 & 98.0 & 99.0 & 98.0 & 97.0 & 97.0 & 92.9 & 93.0 & 97.5 & 96.2 & 97.6 & 98.2 & 95.9 \\ \midrule
p=0.1 & \textbf{98.9} & \textbf{100.0} & \textbf{98.0} & \multicolumn{1}{c}{99.0} & \textbf{96.7} & 99.0 & \textbf{100.0} & \textbf{96.0} & \textbf{99.0} & \textbf{100.0} & \textbf{100.0} & \textbf{98.5} & \textbf{100.0} & \textbf{99.0} & \textbf{100.0} & \textbf{99.0} & \textbf{99.2} & \textbf{98.6} & \textbf{99.8} & \textbf{98.3} \\
p=0.3 & \textbf{98.9} & \textbf{100.0} & \textbf{98.0} & \multicolumn{1}{c}{99.0} & 96.3 & 99.0 & \textbf{100.0} & \textbf{96.0} & \textbf{99.0} & 99.0 & \textbf{100.0} & \textbf{98.5} & \textbf{100.0} & \textbf{99.0} & \textbf{100.0} & \textbf{99.0} & 99.1 & 98.5 & 99.7 & \textbf{98.3} \\
p=0.5 & 98.8 & \textbf{100.0} & \textbf{98.0} & \multicolumn{1}{c}{99.0} & 96.3 & 98.7 & \textbf{100.0} & \textbf{96.0} & \textbf{99.0} & 99.0 & \textbf{100.0} & \textbf{98.5} & \textbf{100.0} & \textbf{99.0} & \textbf{100.0} & 98.5 & 99.1 & 98.5 & 99.6 & \textbf{98.3} \\
p=0.75 & 98.6 & \textbf{100.0} & \textbf{98.0} & \multicolumn{1}{c}{\textbf{99.5}} & 95.7 & \textbf{99.3} & \textbf{100.0} & 92.0 & \textbf{99.0} & \textbf{100.0} & \textbf{100.0} & \textbf{98.5} & \textbf{100.0} & \textbf{99.0} & 99.0 & 98.5 & 98.6 & \textbf{98.6} & 99.7 & 97.8 \\

\bottomrule\toprule
\multicolumn{1}{l}{} & \multicolumn{20}{c}{\textbf{Cross-lingual}} \\
 & \multicolumn{1}{c}{\textbf{avg}} & \multicolumn{1}{c}{\textbf{ar}} & \multicolumn{1}{c}{\textbf{de}} & \textbf{en} & \multicolumn{1}{c}{\textbf{es}} & \multicolumn{1}{c}{\textbf{fr}} & \multicolumn{1}{c}{\textbf{hi}} & \multicolumn{1}{c}{\textbf{id}} & \multicolumn{1}{c}{\textbf{it}} & \multicolumn{1}{c}{\textbf{ja}} & \multicolumn{1}{c}{\textbf{ko}} & \multicolumn{1}{c}{\textbf{pt}} & \multicolumn{1}{c}{\textbf{ru}} & \multicolumn{1}{c}{\textbf{tr}} & \multicolumn{1}{c}{\textbf{vi}} & \multicolumn{1}{c}{\textbf{zh}} & \textbf{$\neg$ IE} & \textbf{IE} & \textbf{$\neg$ Latin} & \textbf{Latin} \\ \midrule
T=0.0 & \textbf{68.1} & 61.7 & 61.9 & - & 70.7 & \textbf{74.6} & 65.2 & 70.4 & 65.9 & \textbf{68.1} & \textbf{70.8} & \textbf{68.8} & 68.8 & 65.6 & \textbf{65.7} & 74.8 & \textbf{68.2} & 67.8 & \textbf{68.2} & 67.9 \\
T=0.3 & \textbf{68.1} & 61.6 & \textbf{63.2} & - & \textbf{72.5} & 74.4 & 65.5 & \textbf{70.8} & 65.7 & 65.3 & 69.2 & 67.2 & \textbf{69.4} & \textbf{67.7} & \textbf{65.7} & 75.0 & 68.1 & 68.1 & 67.7 & \textbf{68.4} \\
T=0.5 & 67.9 & 60.7 & 62.5 & - & 72.1 & 74.3 & 64.3 & 69.6 & \textbf{66.2} & 67.1 & 70.2 & 67.8 & 67.1 & 67.0 & 65.6 & \textbf{76.2} & 67.9 & 67.9 & 67.6 & 68.1 \\
T=0.7 & 67.9 & 60.7 & \textbf{63.2} & - & 72.1 & 74.3 & \textbf{65.6} & 68.8 & \textbf{66.2} & \textbf{68.1} & 69.7 & 68.4 & 68.4 & 64.4 & \textbf{65.7} & 75.5 & 67.7 & \textbf{68.3} & 68.0 & 67.9 \\
T=1.0 & 64.0 & \textbf{63.3} & 63.0 & - & 70.6 & 72.4 & 63.3 & 63.9 & 62.0 & 65.4 & 65.2 & 66.1 & 62.7 & 57.1 & 51.7 & 69.8 & 62.4 & 66.2 & 64.9 & 63.3 \\ \midrule
p=0.1 & 68.0 & 61.3 & 62.5 & - & 71.1 & 75.0 & 65.1 & 70.1 & \textbf{66.9} & 67.8 & \textbf{70.2} & 68.4 & 68.3 & 65.9 & 65.7 & 74.2 & 67.9 & 68.2 & 67.8 & 68.2 \\
p=0.3 & 68.1 & 61.3 & 62.5 & - & 70.7 & 74.6 & 65.4 & 69.7 & 66.6 & 67.7 & 69.8 & 68.1 & 68.4 & 67.2 & 65.7 & \textbf{75.2} & 68.1 & 68.0 & 68.0 & 68.1 \\
p=0.5 & \textbf{68.2} & 61.1 & 62.9 & - & 71.4 & \textbf{75.3} & 65.1 & 69.2 & 66.6 & \textbf{68.1} & 70.1 & \textbf{68.8} & \textbf{69.5} & 66.8 & \textbf{65.9} & 74.5 & \textbf{68.2} & \textbf{68.3} & \textbf{68.1} & \textbf{68.4} \\
p=0.75 & 68.1 & \textbf{61.6} & \textbf{63.2} & - & \textbf{72.5} & 74.4 & \textbf{65.5} & \textbf{70.8} & 65.7 & 65.3 & 69.2 & 67.2 & 69.4 & \textbf{67.7} & 65.7 & 75.0 & 68.1 & 68.1 & 67.7 & \textbf{68.4} \\
 \bottomrule
\end{tabular}%
}
\caption{\textbf{Effect of varying temperature ($T$) or nucleus size ($p$) on \textit{monolingual} and \textit{crosslingual} line-level language confusion (LPR) of \textit{Command R}}. Default values are $p=0.75$ and $T=0.3$. \textbf{Best score}. \textit{($\neg$) IE = (non-) Indo-European language. ($\neg$) Latin = (non-) Latin script.}}
\label{tab:app-vary-pt-lpr}
\end{table*}

\begin{table*}[htb]
\centering
\resizebox{\textwidth}{!}{%
\begin{tabular}{@{}rarrcrrrrrrrrrrrr|cccc@{}}
\toprule
\multicolumn{1}{l}{} & \multicolumn{20}{c}{\textbf{Monolingual}} \\
 & \multicolumn{1}{c}{\textbf{avg}} & \multicolumn{1}{c}{\textbf{ar}} & \multicolumn{1}{c}{\textbf{de}} & \textbf{en} & \multicolumn{1}{c}{\textbf{es}} & \multicolumn{1}{c}{\textbf{fr}} & \multicolumn{1}{c}{\textbf{hi}} & \multicolumn{1}{c}{\textbf{id}} & \multicolumn{1}{c}{\textbf{it}} & \multicolumn{1}{c}{\textbf{ja}} & \multicolumn{1}{c}{\textbf{ko}} & \multicolumn{1}{c}{\textbf{pt}} & \multicolumn{1}{c}{\textbf{ru}} & \multicolumn{1}{c}{\textbf{tr}} & \multicolumn{1}{c}{\textbf{vi}} & \multicolumn{1}{c}{\textbf{zh}} & \textbf{$\neg$ IE} & \textbf{IE} & \textbf{$\neg$ Latin} & \textbf{Latin} \\ \midrule
1 & \textbf{99.0} & \textbf{100.0} & \textbf{98.0} & \textbf{99.5} & \textbf{96.0} & 99.3 & \textbf{100.0} & \textbf{97.0} & 99.0 & \textbf{100.0} & \textbf{100.0} & \textbf{99.5} & \textbf{100.0} & 97.9 & \textbf{100.0} & \textbf{99.0} & \textbf{99.2} & 98.8 & \textbf{99.8} & 98.5 \\
2 & \textbf{99.0} & 99.6 & \textbf{98.0} & \textbf{99.5} & 95.6 & \textbf{99.7} & \textbf{100.0} & 96.0 & 99.0 & \textbf{100.0} & \textbf{100.0} & 99.0 & \textbf{100.0} & \textbf{98.9} & \textbf{100.0} & \textbf{99.0} & \textbf{99.2} & 98.7 & \textbf{99.8} & 98.4 \\
3 & 98.7 & 99.6 & \textbf{98.0} & 99.0 & 95.7 & \textbf{99.7} & 99.0 & 96.0 & 99.0 & 99.0 & \textbf{100.0} & 99.0 & \textbf{100.0} & \textbf{98.9} & \textbf{100.0} & 98.0 & 98.9 & 98.5 & 99.3 & 98.3 \\
5 & \textbf{99.0} & 99.3 & \textbf{98.0} & 99.5 & \textbf{96.0} & \textbf{99.7} & \textbf{100.0} & 96.0 & \textbf{100.0} & \textbf{100.0} & \textbf{100.0} & \textbf{99.5} & \textbf{100.0} & \textbf{98.9} & \textbf{100.0} & 98.5 & 99.1 & \textbf{98.9} & 99.6 & \textbf{98.6} \\
10 & 98.5 & 98.9 & \textbf{98.0} & 99.5 & 95.7 & 99.3 & \textbf{100.0} & 96.0 & \textbf{100.0} & 99.0 & \textbf{100.0} & 98.0 & \textbf{100.0} & 97.8 & 99.0 & 96.9 & 98.4 & 98.6 & 99.1 & 98.1 \\
\bottomrule\toprule
\multicolumn{1}{l}{} & \multicolumn{20}{c}{\textbf{Cross-lingual}} \\
 & \multicolumn{1}{c}{\textbf{avg}} & \multicolumn{1}{c}{\textbf{ar}} & \multicolumn{1}{c}{\textbf{de}} & \textbf{en} & \multicolumn{1}{c}{\textbf{es}} & \multicolumn{1}{c}{\textbf{fr}} & \multicolumn{1}{c}{\textbf{hi}} & \multicolumn{1}{c}{\textbf{id}} & \multicolumn{1}{c}{\textbf{it}} & \multicolumn{1}{c}{\textbf{ja}} & \multicolumn{1}{c}{\textbf{ko}} & \multicolumn{1}{c}{\textbf{pt}} & \multicolumn{1}{c}{\textbf{ru}} & \multicolumn{1}{c}{\textbf{tr}} & \multicolumn{1}{c}{\textbf{vi}} & \multicolumn{1}{c}{\textbf{zh}} & \textbf{$\neg$ IE} & \textbf{IE} & \textbf{$\neg$ Latin} & \textbf{Latin} \\ \midrule
1 & \textbf{74.1} & \textbf{68.4} & \textbf{70.9} & - & \textbf{76.9} & \textbf{77.2} & \textbf{74.7} & \textbf{74.9} & \textbf{72.0} & \textbf{73.5} & \textbf{76.4} & \textbf{73.7} & \textbf{73.6} & \textbf{73.8} & \textbf{73.0} & \textbf{77.8} & \textbf{73.9} & \textbf{74.2} & \textbf{74.1} & \textbf{74.0} \\
2 & 72.2 & 67.0 & 70.5 & - & 76.5 & 75.7 & 74.0 & 72.5 & 70.8 & 70.9 & \textbf{73.8} & \textbf{74.3} & 70.8 & 69.5 & 70.8 & 73.5 & 71.1 & 73.6 & 71.6 & 72.6 \\
3 & 71.5 & 67.2 & 69.8 & - & 76.5 & 76.4 & 75.0 & 69.8 & 71.6 & 69.2 & 73.1 & 71.1 & 69.3 & 67.6 & 70.1 & 74.8 & 70.1 & 73.4 & 71.4 & 71.6 \\
5 & 70.3 & 65.0 & 69.6 & - & 76.2 & 76.3 & 75.1 & 67.0 & 71.0 & 68.2 & 72.4 & 69.1 & 67.6 & 65.5 & 66.7 & 73.8 & 68.3 & 72.9 & 70.4 & 70.2 \\
10 & 68.4 & 65.0 & 68.6 & - & 75.9 & \textbf{75.7} & \textbf{75.7} & 66.2 & 70.2 & 66.1 & 68.8 & 66.7 & 63.7 & 60.6 & 64.7 & 69.6 & 65.6 & 72.1 & 68.1 & 68.6 \\
 \bottomrule
\end{tabular}%
}
\caption{\textbf{\textit{Monolingual} and \textit{Cross-lingual} line-level pass rate (LPR) of \textit{Command R} using beam search} with beam sizes 1-10. \textit{($\neg$) IE = (non-) Indo-European language. ($\neg$) Latin = (non-) Latin script.}}
\label{tab:beam-search-full-lpr}
\end{table*}

\clearpage
\newpage
\onecolumn
\subsection{Base vs Instruction Tuning comparison}
\label{sec:base_vs_instruct}

\begin{figure*}[h]
\begin{tcolorbox}[width=\linewidth, sharp corners=all, colback=white!95!black]
\textbf{Q:} Write your answer in French. How should I choose what cheese to buy?

\textbf{A:} Il existe de nombreux types de fromages différents, donc le choix du fromage à acheter dépend des préférences personnelles, de la disponibilité et de l'utilisation prévue. [...]

\textbf{Q:} What is the difference between pets and cattle? Reply in Arabic.

\textbf{A:}
\end{tcolorbox} 
\caption{Template used for few-shot prompting the base models. The model's answers are truncated to prevent the generation of new questions. For the instruct variants, we use similar prompting, except that the Q/A examples are formatted as User/Chatbot turns using the model's chat template.}
\label{fig:few_shot_template}
\end{figure*}

\begin{table*}[ht]
\centering
\resizebox{\textwidth}{!}{%
\begin{tabular}{larrrrrrrrrrrrrrr}
\toprule
 & Avg & ar & de & en & es & fr & hi & id & it & ja & ko & pt & ru & tr & vi & zh \\ \midrule
Llama 2 70B     & \textbf{97.5}        & \textbf{100.0}            & \textbf{92.9}        & \textbf{99.5}                 & \textbf{98.0}        & \textbf{96.5}        & \textbf{98.9}        & \textbf{91.9}        & \textbf{98.0}        & \textbf{100.0}       & \textbf{100.0}       & \textbf{95.8}        & \textbf{100.0}       & \textbf{98.7}        & \textbf{98.0}        & \textbf{93.6}        \\
Llama 2 70B-I   & 48.3          & 0.3            & 59.0           & 99.0  & 95.7          & 87.7          & 1.0            & 62.0          & 72.0           & 7.0            & 0.0            & 91.0          & 88.9           & 33.0           & 17.0          & 10.5          \\\midrule
Llama 3 70B     & \textbf{94.8}        & \textbf{95.6}             & \textbf{96.9}        & 99.0                 & 96.6                 & \textbf{97.6}        & \textbf{97.9}        & \textbf{82.8}        & \textbf{93.9}        & \textbf{88.9}        & \textbf{100.0}       & \textbf{96.2}        & \textbf{100.0}       & \textbf{94.4}        & \textbf{93.9}        & \textbf{89.1}        \\
Llama 3 70B-I   & 46.0          & 21.7           & 31.0           & \textbf{100.0} & \textbf{98.3} & 88.7          & 23.0           & 21.0          & 88.0           & 10.0           & 0.0            & 95.5          & 77.0           & 18.0           & 10.0          & 8.0           \\\midrule
Command R base  & 86.3          & 94.5           & 83.8           & 98.5           & 89.6          & 86.6          & 81.0           & 69.4          & 79.0           & 98.0           & 94.3           & 83.9          & 93.6           & 91.0           & 79.6          & 71.1          \\
Command R       & \textbf{98.6} & \textbf{100.0} & \textbf{98.0}  & \textbf{99.5}  & \textbf{95.7} & \textbf{99.3} & \textbf{100.0} & \textbf{92.0} & \textbf{99.0}  & \textbf{100.0} & \textbf{100.0} & \textbf{98.5} & \textbf{100.0} & \textbf{99.0}  & \textbf{99.0} & \textbf{98.5} \\\midrule
Command R+ base & 82.1          & 93.2           & 71.4           & 98.0           & 81.9          & 86.7          & 68.7           & 80.8          & 65.0           & 92.7           & 92.1           & 79.8          & 93.5           & 95.0           & 70.7          & 62.4          \\
Command R+      & \textbf{99.2} & \textbf{99.7}  & \textbf{100.0} & \textbf{100.0} & \textbf{99.3} & \textbf{99.7} & \textbf{100.0} & \textbf{97.0} & \textbf{100.0} & \textbf{99.0}  & \textbf{100.0} & \textbf{97.5} & \textbf{100.0} & \textbf{100.0} & \textbf{99.0} & \textbf{97.5} \\
\bottomrule
\end{tabular}%
}
\caption{\textbf{LPR of base vs instruction-tuned LLMs for \textit{monolingual} generation.}}
\label{tab:impact-instruction-tuning-monolingual}
\end{table*}
\
\begin{table*}[h]
\centering
\resizebox{\columnwidth}{!}{%
\setlength\tabcolsep{3.5pt} 
\begin{tabular}{l|c|cccccc|c|cccccccccccccc}
\toprule
\multirow{2}{*}{Model} & \multicolumn{7}{c|}{Crosslingual WPR} & \multicolumn{15}{c}{Crosslingual LPR} \\
& avg & ar & hi & ja & ko & ru & zh & avg & ar & de & es & fr & hi & id & it & ja & ko & pt & ru & tr & vi & zh \\
\hline
Command R Base & 100.0 & 100.0 & 100.0 & 100.0 & 100.0 & 100.0 & 100.0 & 1.1 & 0.0 & 0.9 & 1.2 & 1.6 & 0.8 & 1.4 & 0.3 & 2.4 & 0.0 & 0.9 & 0.0 & 0.4 & 1.6 & 3.6 \\
+ Q/A template & 97.0 & 95.2 & 100.0 & 99.0 & 97.9 & 91.7 & 98.4 & 20.9 & 3.3 & 19.9 & 27.6 & 27.5 & 5.3 & 25.8 & 23.2 & 42.5 & 13.5 & 17.5 & 8.0 & 19.6 & 14.9 & 44.0 \\
+ 1-shot & 98.6 & 98.5 & 98.8 & 98.6 & 98.3 & 98.4 & 99.2 & 90.7 & 83.8 & 93.3 & 95.0 & 93.2 & 84.2 & 86.7 & 93.0 & 93.2 & 92.5 & 91.1 & 88.3 & 89.6 & 91.9 & 93.3 \\
+ 5-shot & 99.7 & 100.0 & 99.2 & 100.0 & 99.5 & 100.0 & 99.6 & 95.0 & 91.0 & 96.3 & 96.8 & 95.9 & 96.0 & 92.8 & 96.5 & 93.8 & 95.5 & 95.4 & 93.8 & 96.8 & 95.6 & 94.5 \\
+ English SFT & 91.7 & 95.5 & 98.7 & 77.9 & 94.8 & 94.3 & 89.0 & 78.3 & 70.0 & 77.7 & 82.7 & 76.6 & 78.4 & 77.7 & 74.8 & 80.8 & 80.7 & 79.3 & 74.4 & 82.4 & 83.9 & 77.1 \\
\quad{} + English pref. tuning & 87.4 & 91.2 & 97.7 & 70.4 & 92.9 & 88.5 & 83.6 & 85.7 & 86.4 & 87.4 & 86.2 & 82.1 & 87.7 & 80.2 & 83.9 & 87.6 & 84.3 & 83.2 & 86.7 & 90.1 & 89.6 & 84.6 \\
+ Multilingual SFT & 90.0 & 96.0 & 97.6 & 76.0 & 92.1 & 89.0 & 89.7 & 78.2 & 89.1 & 62.4 & 76.8 & 69.4 & 82.8 & 77.3 & 72.6 & 77.7 & 81.1 & 76.8 & 79.5 & 82.3 & 85.1 & 82.4 \\
\quad{} + Multi. pref. tuning & 86.9 & 94.1 & 96.6 & 71.2 & 87.7 & 84.9 & 87.0 & 89.4 & 94.1 & 80.9 & 91.9 & 88.7 & 91.2 & 85.6 & 88.9 & 89.2 & 93.0 & 83.7 & 92.3 & 91.7 & 93.0 & 87.8 \\
\emph{Command R} & 94.0 & 94.3 & 98.6 & 88.5 & 97.2 & 94.0 & 91.1 & 68.1 & 61.6 & 63.2 & 72.5 & 74.4 & 65.5 & 70.8 & 65.7 & 65.3 & 69.2 & 67.2 & 69.4 & 67.7 & 65.7 & 75.0 \\
+ 1-shot & 92.3 & 97.2 & 98.3 & 87.6 & 89.7 & 91.6 & 89.5 & 82.9 & 80.0 & 79.5 & 85.6 & 82.6 & 84.7 & 79.9 & 82.4 & 78.0 & 87.2 & 81.5 & 84.4 & 84.5 & 84.5 & 85.2 \\

\bottomrule
\end{tabular}
}
\caption{\textbf{Effect of few-shot prompting and instruction tuning on cross-lingual language confusion, detailed per-language results.}}
\label{tab:ablation_crosslingual_detailed}
\end{table*}

\begin{table*}[h]
\centering
\resizebox{\columnwidth}{!}{%
\setlength\tabcolsep{2.5pt} 
\begin{tabular}{l|c|cccccc|c|ccccccccccccccc}
\toprule
\multirow{2}{*}{Model} & \multicolumn{7}{c|}{Monolingual WPR} & \multicolumn{14}{c}{Monolingual LPR} \\
& avg & ar & hi & ja & ko & ru & zh & avg & ar & de & en & es & fr & hi & id & it & ja & ko & pt & ru & tr & vi & zh \\
\hline
Command R Base & 98.7 & 97.9 & 95.1 & 100.0 & 100.0 & 100.0 & 99.4 & 86.2 & 94.9 & 85.0 & 99.5 & 90.5 & 85.3 & 81.0 & 72.2 & 74.0 & 93.9 & 94.2 & 84.0 & 94.8 & 92.0 & 83.0 & 68.1 \\
+ Q/A template & 99.7 & 100.0 & 100.0 & 100.0 & 100.0 & 98.5 & 100.0 & 85.3 & 87.6 & 73.2 & 100.0 & 89.8 & 86.3 & 98.9 & 49.0 & 91.9 & 97.9 & 93.5 & 96.8 & 86.7 & 96.3 & 49.0 & 82.4 \\
+ 1-shot & 100.0 & 100.0 & 100.0 & 100.0 & 100.0 & 100.0 & 100.0 & 94.1 & 93.0 & 99.0 & 100.0 & 99.3 & 47.5 & 96.9 & 93.8 & 99.0 & 99.0 & 96.8 & 99.0 & 97.3 & 95.1 & 96.6 & 99.0 \\
+ 5-shot & 100.0 & 100.0 & 100.0 & 100.0 & 100.0 & 100.0 & 100.0 & 99.0 & 99.6 & 99.0 & 100.0 & 100.0 & 98.3 & 98.0 & 96.9 & 100.0 & 100.0 & 99.0 & 99.5 & 98.9 & 98.8 & 99.0 & 97.9 \\
+ English SFT & 96.2 & 98.4 & 100.0 & 88.1 & 98.5 & 100.0 & 92.3 & 77.8 & 54.1 & 83.8 & 100.0 & 80.3 & 86.9 & 59.6 & 89.0 & 96.0 & 67.7 & 66.7 & 84.4 & 68.0 & 64.0 & 85.0 & 82.0 \\
\quad{} + English pref. tuning & 90.9 & 93.5 & 97.8 & 84.4 & 95.4 & 89.6 & 85.0 & 74.3 & 54.4 & 81.0 & 100.0 & 82.7 & 85.6 & 46.0 & 80.0 & 95.0 & 64.0 & 65.0 & 81.4 & 67.7 & 51.0 & 82.0 & 78.5 \\
+ Multilingual SFT & 95.5 & 97.9 & 100.0 & 86.6 & 97.9 & 99.0 & 91.6 & 98.3 & 99.7 & 100.0 & 100.0 & 99.7 & 98.7 & 100.0 & 90.0 & 99.0 & 98.0 & 98.0 & 98.4 & 100.0 & 96.8 & 100.0 & 97.0 \\
\quad{} + Multi. pref. tuning & 93.4 & 97.9 & 100.0 & 86.7 & 93.0 & 91.8 & 91.2 & 98.8 & 99.6 & 97.0 & 99.5 & 99.3 & 99.3 & 100.0 & 92.0 & 100.0 & 99.0 & 100.0 & 99.0 & 100.0 & 100.0 & 100.0 & 98.0 \\
\emph{Command R} & 96.3 & 99.3 & 99.0 & 93.9 & 97.0 & 96.0 & 92.3 & 98.6 & 100.0 & 98.0 & 99.5 & 95.7 & 99.3 & 100.0 & 92.0 & 99.0 & 100.0 & 100.0 & 98.5 & 100.0 & 99.0 & 99.0 & 98.5 \\
+ 1-shot & 92.7 & 94.4 & 95.7 & 93.8 & 88.9 & 90.5 & 92.9 & 68.3 & 51.8 & 53.0 & 5.5 & 83.2 & 98.3 & 23.7 & 88.0 & 94.0 & 65.0 & 95.7 & 66.0 & 95.5 & 90.3 & 97.8 & 17.0 \\

\bottomrule
\end{tabular}
}
\caption{\textbf{Effect of few-shot prompting and instruction tuning on monolingual language confusion, detailed per-language results.}}
\label{tab:ablation_monolingual_detailed}
\end{table*}

\clearpage
\subsection{Metric Variability} \label{sec:app-metric-variability}
We find there is little variability LPR and WPR between runs.  We run monolingual and crosslingual language confusion tasks on the FP16 and W4g variants of Command R/R+ 5 times each, and report the minimum, maximum, and mean of LPR and WPR in Table \ref{tab:app-lprwpr-variability}.

\begin{table*}[htb]
    \centering
    \resizebox{\textwidth}{!}{%
    \begin{tabular}{@{}ll|rrrrrrrrrrrrrrr|rrrrrrr@{}}
     &  & \multicolumn{15}{c|}{\textbf{Monolingual LPR}} & \multicolumn{7}{c}{\textbf{Monolingual WPR}} \\
     &  & \multicolumn{1}{c|}{\textbf{avg}} & \multicolumn{1}{c}{\textbf{ar}} & \multicolumn{1}{c}{\textbf{de}} & \multicolumn{1}{c}{\textbf{es}} & \multicolumn{1}{c}{\textbf{fr}} & \multicolumn{1}{c}{\textbf{hi}} & \multicolumn{1}{c}{\textbf{id}} & \multicolumn{1}{c}{\textbf{it}} & \multicolumn{1}{c}{\textbf{ja}} & \multicolumn{1}{c}{\textbf{ko}} & \multicolumn{1}{c}{\textbf{pt}} & \multicolumn{1}{c}{\textbf{ru}} & \multicolumn{1}{c}{\textbf{tr}} & \multicolumn{1}{c}{\textbf{vi}} & \multicolumn{1}{c|}{\textbf{zh}} & \multicolumn{1}{c|}{\textbf{avg}} & \multicolumn{1}{c}{\textbf{ar}} & \multicolumn{1}{c}{\textbf{hi}} & \multicolumn{1}{c}{\textbf{ja}} & \multicolumn{1}{c}{\textbf{ko}} & \multicolumn{1}{c}{\textbf{ru}} & \multicolumn{1}{c}{\textbf{zh}} \\ \midrule
    \multicolumn{1}{l|}{Command R+ fp16} & min & \multicolumn{1}{r|}{99.1} & 98.6 & 100.0 & 99.0 & 98.7 & 100.0 & 95.0 & 100.0 & 98.0 & 100.0 & 97.5 & 100.0 & 100.0 & 99.0 & 97.0 & \multicolumn{1}{r|}{99.1} & 99.3 & 100.0 & 99.0 & 98.0 & 97.0 & 100.0 \\
    \multicolumn{1}{l|}{} & mean & \multicolumn{1}{r|}{99.3} & 99.3 & 100.0 & 99.3 & 99.6 & 100.0 & 96.6 & 100.0 & 98.6 & 100.0 & 98.3 & 100.0 & 100.0 & 99.8 & 97.9 & \multicolumn{1}{r|}{99.4} & 99.7 & 100.0 & 99.8 & 99.4 & 97.6 & 100.0 \\
    \multicolumn{1}{l|}{} & max & \multicolumn{1}{r|}{99.5} & 100.0 & 100.0 & 99.7 & 100.0 & 100.0 & 99.0 & 100.0 & 99.0 & 100.0 & 99.0 & 100.0 & 100.0 & 100.0 & 99.0 & \multicolumn{1}{r|}{99.6} & 100.0 & 100.0 & 100.0 & 100.0 & 99.0 & 100.0 \\ \midrule
    \multicolumn{1}{l|}{Command R+ w4g} & min & \multicolumn{1}{r|}{99.3} & 98.6 & 100.0 & 99.3 & 99.7 & 100.0 & 97.0 & 100.0 & 97.0 & 100.0 & 97.5 & 100.0 & 100.0 & 100.0 & 98.5 & \multicolumn{1}{r|}{99.5} & 100.0 & 100.0 & 99.0 & 99.0 & 99.0 & 98.5 \\
    \multicolumn{1}{l|}{} & mean & \multicolumn{1}{r|}{99.4} & 99.1 & 100.0 & 99.6 & 99.9 & 100.0 & 97.8 & 100.0 & 97.4 & 100.0 & 98.1 & 100.0 & 100.0 & 100.0 & 98.9 & \multicolumn{1}{r|}{99.7} & 100.0 & 100.0 & 99.8 & 99.8 & 99.4 & 99.2 \\
    \multicolumn{1}{l|}{} & max & \multicolumn{1}{r|}{99.5} & 99.6 & 100.0 & 99.7 & 100.0 & 100.0 & 99.0 & 100.0 & 98.0 & 100.0 & 98.5 & 100.0 & 100.0 & 100.0 & 99.5 & \multicolumn{1}{r|}{99.8} & 100.0 & 100.0 & 100.0 & 100.0 & 100.0 & 100.0 \\ \midrule
    \multicolumn{1}{l|}{Command R fp16} & min & \multicolumn{1}{r|}{98.3} & 98.6 & 97.0 & 98.0 & 98.7 & 100.0 & 92.0 & 99.0 & 98.0 & 99.0 & 98.5 & 99.0 & 98.0 & 100.0 & 96.5 & \multicolumn{1}{r|}{95.1} & 97.3 & 95.0 & 90.7 & 97.0 & 93.9 & 92.3 \\
    \multicolumn{1}{l|}{} & mean & \multicolumn{1}{r|}{98.6} & 99.2 & 97.0 & 98.1 & 99.2 & 100.0 & 93.6 & 99.6 & 99.6 & 99.0 & 99.0 & 99.8 & 98.6 & 100.0 & 97.7 & \multicolumn{1}{r|}{95.6} & 97.8 & 96.0 & 92.5 & 97.6 & 96.2 & 93.4 \\
    \multicolumn{1}{l|}{} & max & \multicolumn{1}{r|}{98.8} & 99.3 & 97.0 & 98.3 & 99.3 & 100.0 & 95.0 & 100.0 & 100.0 & 99.0 & 99.5 & 100.0 & 99.0 & 100.0 & 98.5 & \multicolumn{1}{r|}{96.5} & 98.7 & 97.0 & 96.0 & 98.0 & 97.0 & 94.4 \\ \midrule
    \multicolumn{1}{l|}{Command R w4g} & min & \multicolumn{1}{r|}{98.5} & 99.0 & 95.0 & 96.3 & 99.7 & 99.0 & 96.0 & 100.0 & 99.0 & 97.0 & 98.0 & 100.0 & 98.0 & 99.0 & 98.0 & \multicolumn{1}{r|}{95.7} & 98.3 & 97.0 & 86.9 & 95.9 & 96.0 & 93.9 \\
    \multicolumn{1}{l|}{} & mean & \multicolumn{1}{r|}{98.6} & 99.4 & 95.0 & 96.5 & 99.9 & 99.8 & 97.2 & 100.0 & 99.8 & 97.0 & 98.3 & 100.0 & 98.8 & 99.8 & 98.6 & \multicolumn{1}{r|}{95.9} & 98.4 & 97.6 & 89.1 & 96.7 & 98.4 & 95.2 \\
    \multicolumn{1}{l|}{} & max & \multicolumn{1}{r|}{98.7} & 100.0 & 95.0 & 97.0 & 100.0 & 100.0 & 98.0 & 100.0 & 100.0 & 97.0 & 98.5 & 100.0 & 99.0 & 100.0 & 99.5 & \multicolumn{1}{r|}{96.2} & 98.6 & 98.0 & 91.9 & 97.9 & 100.0 & 96.0 \\ \midrule \bottomrule
    
    &  &  &  &  &  &  &  &  &  &  &  &  &  &  &  &  &  &  &  &  &  &  &  \\
    
     &  & \multicolumn{15}{c|}{\textbf{Crosslingual LPR}} & \multicolumn{7}{c}{\textbf{Crosslingual WPR}} \\
     &  & \multicolumn{1}{c|}{\textbf{avg}} & \multicolumn{1}{c}{\textbf{ar}} & \multicolumn{1}{c}{\textbf{de}} & \multicolumn{1}{c}{\textbf{es}} & \multicolumn{1}{c}{\textbf{fr}} & \multicolumn{1}{c}{\textbf{hi}} & \multicolumn{1}{c}{\textbf{id}} & \multicolumn{1}{c}{\textbf{it}} & \multicolumn{1}{c}{\textbf{ja}} & \multicolumn{1}{c}{\textbf{ko}} & \multicolumn{1}{c}{\textbf{pt}} & \multicolumn{1}{c}{\textbf{ru}} & \multicolumn{1}{c}{\textbf{tr}} & \multicolumn{1}{c}{\textbf{vi}} & \multicolumn{1}{c|}{\textbf{zh}} & \multicolumn{1}{c|}{\textbf{avg}} & \multicolumn{1}{c}{\textbf{ar}} & \multicolumn{1}{c}{\textbf{hi}} & \multicolumn{1}{c}{\textbf{ja}} & \multicolumn{1}{c}{\textbf{ko}} & \multicolumn{1}{c}{\textbf{ru}} & \multicolumn{1}{c}{\textbf{zh}} \\ \midrule
    \multicolumn{1}{l|}{Command R+ fp16} & min & \multicolumn{1}{r|}{90.8} & 92.5 & 90.3 & 90.0 & 91.1 & 88.1 & 87.8 & 92.4 & 92.4 & 90.6 & 87.5 & 89.8 & 91.5 & 90.1 & 90.2 & \multicolumn{1}{r|}{94.8} & 97.5 & 94.4 & 94.7 & 94.6 & 90.2 & 94.4 \\
    \multicolumn{1}{l|}{} & mean & \multicolumn{1}{r|}{91.0} & 93.0 & 90.6 & 91.2 & 91.6 & 88.8 & 88.2 & 93.0 & 93.1 & 91.1 & 88.3 & 91.3 & 92.0 & 90.8 & 91.3 & \multicolumn{1}{r|}{95.1} & 97.9 & 95.2 & 95.7 & 95.1 & 91.1 & 95.9 \\
    \multicolumn{1}{l|}{} & max & \multicolumn{1}{r|}{91.2} & 93.6 & 91.0 & 92.4 & 91.9 & 89.2 & 88.6 & 93.4 & 94.1 & 91.3 & 89.6 & 92.4 & 92.9 & 91.5 & 92.5 & \multicolumn{1}{r|}{95.4} & 98.3 & 95.7 & 96.6 & 95.6 & 92.0 & 96.5 \\ \midrule
    \multicolumn{1}{l|}{Command R+ w4g} & min & \multicolumn{1}{r|}{90.0} & 89.9 & 89.2 & 90.4 & 91.6 & 87.5 & 85.6 & 92.5 & 92.4 & 89.8 & 84.1 & 90.6 & 89.8 & 90.4 & 89.2 & \multicolumn{1}{r|}{94.8} & 97.2 & 94.2 & 96.0 & 94.9 & 88.7 & 95.7 \\
    \multicolumn{1}{l|}{} & mean & \multicolumn{1}{r|}{90.1} & 90.6 & 89.9 & 90.7 & 91.7 & 88.4 & 86.5 & 93.1 & 92.8 & 90.6 & 85.4 & 91.0 & 90.7 & 90.7 & 89.6 & \multicolumn{1}{r|}{95.3} & 97.7 & 94.8 & 96.5 & 95.8 & 90.3 & 96.5 \\
    \multicolumn{1}{l|}{} & max & \multicolumn{1}{r|}{90.2} & 91.1 & 90.7 & 91.0 & 91.7 & 89.1 & 87.4 & 93.8 & 93.4 & 91.1 & 85.9 & 91.4 & 91.5 & 91.0 & 90.5 & \multicolumn{1}{r|}{95.6} & 98.2 & 95.9 & 97.3 & 96.4 & 91.1 & 97.6 \\ \midrule
    \multicolumn{1}{l|}{Command R fp16} & min & \multicolumn{1}{r|}{66.3} & 57.9 & 58.6 & 68.8 & 72.1 & 63.9 & 68.7 & 62.9 & 65.1 & 68.6 & 63.8 & 67.4 & 63.6 & 64.2 & 74.2 & \multicolumn{1}{r|}{92.9} & 97.5 & 97.6 & 88.6 & 90.5 & 86.6 & 90.2 \\
    \multicolumn{1}{l|}{} & mean & \multicolumn{1}{r|}{66.4} & 58.8 & 59.6 & 69.0 & 73.0 & 64.0 & 69.3 & 63.6 & 66.3 & 69.2 & 64.2 & 68.8 & 64.2 & 65.1 & 74.6 & \multicolumn{1}{r|}{93.6} & 98.5 & 98.4 & 90.5 & 93.3 & 88.7 & 92.1 \\
    \multicolumn{1}{l|}{} & max & \multicolumn{1}{r|}{66.6} & 59.5 & 60.0 & 69.5 & 73.8 & 64.3 & 69.8 & 63.9 & 67.4 & 69.6 & 64.4 & 69.7 & 64.8 & 66.3 & 74.8 & \multicolumn{1}{r|}{94.4} & 99.2 & 99.6 & 92.1 & 95.1 & 91.0 & 93.4 \\ \midrule
    \multicolumn{1}{l|}{Command R w4g} & min & \multicolumn{1}{r|}{67.0} & 60.3 & 58.9 & 72.4 & 73.0 & 67.3 & 66.6 & 64.2 & 65.0 & 69.5 & 64.1 & 68.9 & 64.2 & 65.7 & 72.5 & \multicolumn{1}{r|}{92.5} & 96.3 & 94.8 & 89.2 & 92.8 & 86.5 & 90.8 \\
    \multicolumn{1}{l|}{} & mean & \multicolumn{1}{r|}{67.2} & 60.4 & 59.3 & 72.8 & 73.3 & 67.7 & 67.9 & 64.8 & 65.4 & 70.4 & 64.6 & 69.7 & 64.6 & 66.1 & 73.0 & \multicolumn{1}{r|}{93.0} & 97.5 & 95.6 & 90.3 & 94.8 & 88.0 & 92.3 \\
    \multicolumn{1}{l|}{} & max & \multicolumn{1}{r|}{67.3} & 60.6 & 59.6 & 73.0 & 73.8 & 68.0 & 68.6 & 65.2 & 66.0 & 71.1 & 65.1 & 70.6 & 65.2 & 66.4 & 73.8 & \multicolumn{1}{r|}{93.6} & 98.4 & 96.2 & 91.6 & 95.9 & 89.3 & 93.6 \\ \bottomrule
    \end{tabular}%
    }
\caption{\textbf{Variability in LPR \& WPR metrics for Command R/R+ models over 5 evaluation runs.}}
\label{tab:app-lprwpr-variability}
\end{table*}

\clearpage
\newpage
\twocolumn
\subsection{Nucleus Sampling with Temperature}
\label{sec:app-nuc-sampling}
Previous work shows that greedy search empirically leads to repetition and generally low-quality generations.  One solution is adding stochasticity to decoding, such as by sampling the next token from the top-K next most likely tokens or from the top-P of the probability distribution over next tokens. The latter is called ``nucleus sampling'' \cite{holtzman2019curious}, and can be combined with top-K and temperature sampling.  

Let $\boldsymbol{x} = [x_0, x_1, x_2, ..., x_{n-1}]$ be a sequence of tokens.  We describe how to sample the next token in the sequence via \textit{nucleus sampling with temperature} given a language model with vocabulary $V$.

\begin{figure}[h]
    \centering
    \includegraphics[width=1\linewidth]{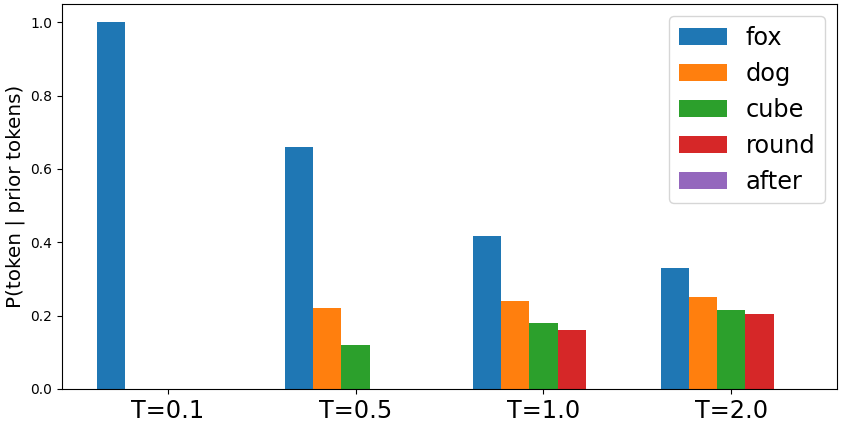}
    \caption{\textbf{Effect of increasing temperature $T$ on output distribution.} As $T$ $\uparrow$, the distribution flattens.}
    \label{fig:flattened_dist}
\end{figure}

The \textit{nucleus} of a probability distribution for $0~<~p~\leq~1$ is the smallest subset $V'~\subset~V$ such that the summed probabilities of $V'$ are greater than or equal to $p$.\footnote{Some implementations will define $V'$ as the \textit{largest subset} $V'~\subset~V$ s.t. the summed probabilities of $V'$ is \textit{less-than-or-equal-to} $p$.  We use the definition from the original work.}
\textit{Nucleus sampling} samples the next token $x_n$ at random from this set.  The function \textit{top\_p} returns $V'$:
\begin{equation*}
\label{eq:get_nucleus}
    \text{top\_p}(\boldsymbol{x}, V, p) = v \in V \text{ s.t. } \sum_i P(v_i | \boldsymbol{x}) \geq p
\end{equation*}
\noindent Let $z \in \mathbb{R}^{|V|}$ be an output vector of logits.  We apply the softmax function over element $z_i$ given temperature $T$ to transform its logit $z_i$ to a probability over vocabulary tokens, as below. Let $v_i$ be the vocabulary token corresponding to logit $z_i$. 
\begin{equation}
\label{eq:softmax}
P(v_i | \boldsymbol{x}) = \textit{softmax}(z_i, T) = \frac{e^\frac{z_i}{T}}{\sum_{j}e^\frac{z_j}{T}}
\end{equation}
\noindent The next token $x_n$ is chosen at random from normalized probabilities in the nucleus:
\begin{equation*}
\label{eq:normed_nucleus}
P_{normalized}(v'_i | \boldsymbol{x}) = \frac{P(v'_i | \boldsymbol{x})}{\sum_{v'_j \in V'}P(v'_j | \boldsymbol{x})}
\end{equation*}
\noindent$T$ controls the peakiness of the distribution at the sampling point.  As $T$ increases, the distribution becomes more uniform.  Figure \ref{fig:flattened_dist} shows the effect on Equation \ref{eq:softmax} for a toy example.  As $T$ increases, previously unlikely words become more likely, and highly likely words have lower probabilities. 

We walk through Figure \ref{fig:nucleus-sampling} as an example.  

\paragraph{Lowering Temperature} Imagine we run inference with $p=0.75, T=1$. Given previous tokens \textit{`the'}, \textit{` quick'}, and \textit{` brown'}, the next token options are \textit{` fox'}, \textit{` dog'}, \textit{` cube'}, \begin{CJK*}{UTF8}{gbsn} ` 狐狸'\end{CJK*}, and \textit{` after'} with logits $[0.75, 0.20, -0.10, -0.20, -0.30]$.  Applying the softmax, \textit{` fox'}, \textit{` dog'}, \textit{` cube'}, and \begin{CJK*}{UTF8}{gbsn} ` 狐狸'\end{CJK*} remain in the nucleus with respective likelihoods $[0.418, 0.241, 0.179, 0.162]$.  ` \begin{CJK*}{UTF8}{gbsn} 狐狸\end{CJK*}' has an over 16\% chance of being sampled at the next step---a high risk for language confusion. 
Increasing $T$ to $2.0$, ` \begin{CJK*}{UTF8}{gbsn} 狐狸\end{CJK*}' has an over 20\% chance of being elicited.  Reducing $T$ to $0.5$, however, we sharpen the distribution by shifting probability mass to \textit{` fox'}, \textit{` dog'}, and \textit{` cube'} such that ` \begin{CJK*}{UTF8}{gbsn} 狐狸\end{CJK*}' falls out of the nucleus and cannot be sampled, thus eliminating the risk of language confusion at this timestep.

\paragraph{Reducing Nucleus Size} The distribution may also be sharpened by decreasing nucleus size.  Keeping $T=1$ but reducing nucleus size to $p=0.7$, ` \begin{CJK*}{UTF8}{gbsn} 狐狸\end{CJK*}' is not in the nucleus. The probabilities over next tokens in the nucleus are then: 
${\textit{`` fox''}: 0.499, \textit{`` dog''}: 0.287, \textit{`` cube''}: 0.213}$, and the risk of language confusion is eliminated.   

\newpage

\subsection{Discussion} \label{app:discussion}

\paragraph{Behavior of base models} As observed in \S\ref{sec:impact-instruction-tuning}, language confusion in base models is not correlated with their downstream performance. Stronger base models such as Command R+ and Llama 3 70B are more confused than Command R and Llama 2 70B respectively. Given the occurrence of translations and token in other languages in pre-training \cite{blevins-zettlemoyer-2022-language}, we expect base models to exhibit some degree of language switching while instruction tuning then reinforces the desired behavior.

\paragraph{English-centricity} Despite LLMs in general exhibiting impressive multilingual generative capabilities, the fact that they are most likely to switch to English---both on the sentence and word level---is another example of their English-centric nature \cite{hu2020xtreme,zhao2024llama}. Our experiments in \S\ref{sec:mitigation-instruction-tuning} further highlight the negative impact of overly English-centric instruction tuning, which is also illustrated by Llama Instruct models' high language confusion.

\paragraph{Preference Tuning}  In Table \ref{tab:ablation}, we observe that WPR decreases after preference tuning.  Citing similar observations by \citet{yuan2024advancingllmreasoninggeneralists},\footnote{Also, \url{https://wandb.ai/eric_anthony_mitchell/dpo-demos/runs/og8q3euz?nw=nwusereric_anthony_mitchell}} \citet{yan20243dpropertiesidentifyingchallengesdpo} observe a decrease in the likelihood of both preferred and unpreferred data points as DPO \citep{rafailov_dpo} training progresses, and propose a theoretical explanation. \citet{xu2024dposuperiorppollm} remark that ``DPO is prone to generating a biased policy that favors out-of-distribution responses, leading to unpredictable behaviors.''  It is plausible that if preference learning decreases token likelihoods for examples seen during its training (e.g. common English words), then the relative likelihoods of unseen/rare tokens increases, explaining the large decrease in WPR we see for +English SFT +English pref. tuning. The hypothesis that preference learning encourages unfavorable behaviors such as language confusion is worthy of further exploration.  

\paragraph{Other factors} There are other factors that may affect language confusion. We suspect that word-level confusion is related to under-training. Certain non-English tokens, particularly in low-density regions that are rarely encountered during pre-training, may be under-trained and lack calibration. This can then lead to English tokens being sampled due to the absence of in-language tokens with higher likelihood as seen in \S\ref{sec:when-does-language-confusion-occur}. Future work might investigate this hypothesis.

\end{document}